\documentclass[letterpaper, 10 pt, conference]{ieeeconf}  

\IEEEoverridecommandlockouts                              

\usepackage{amsmath} 
\usepackage{amssymb}  
\usepackage{cite}
\usepackage{amsmath,amssymb,amsfonts}
\usepackage{algorithmic}
\usepackage{graphicx}
\usepackage{textcomp}
\usepackage{subcaption}
\usepackage{xcolor}
\usepackage{booktabs}
\usepackage{float}
\usepackage{nicematrix}
\usepackage{url}
\usepackage{placeins}
\title{\LARGE \bf
SPLATART: Articulated Gaussian Splatting with Estimated Object Structure}

\author{Stanley Lewis$^{1}$, Vishal Chandra$^{1}$, Tom Gao$^{1}$, and Odest Chadwicke Jenkins$^{1}$
\thanks{$^{1}$S. Lewis, Vishal Chandra, Tom Gao, and O.C. Jenkins are with the Robotics Department,
        University of Michigan, Ann Arbor, MI 48109 \{{\tt\small stanlew, chandrav, zimingg, ocj\}@umich.edu}\newline This work is supported in part by Ford Motor Company, in part by J.P. Morgan AI Research, and in part by Amazon}}%

\begin{document}

\maketitle
\thispagestyle{empty}
\pagestyle{empty}

\begin{abstract}



Representing articulated objects remains a difficult problem within the field of robotics. Objects such as pliers, clamps, or cabinets require representations that capture not only geometry and color information, but also part seperation, connectivity, and joint parametrization. Furthermore, learning these representations becomes even more difficult with each additional degree of freedom. Complex articulated objects such as robot arms may have seven or more degrees of freedom, and the depth of their kinematic tree may be notably greater than the tools, drawers, and cabinets that are the typical subjects of articulated object research. To address these concerns, we introduce SPLATART - a pipeline for learning Gaussian splat representations of articulated objects from posed images, of which a subset contains image space part segmentations. SPLATART disentangles the part separation task from the articulation estimation task, allowing for post-facto determination of joint estimation and representation of articulated objects with deeper kinematic trees than previously exhibited. In this work, we present data on the SPLATART pipeline as applied to the syntheic Paris dataset objects \cite{liu2023paris}, and qualitative results on a real-world object under spare segmentation supervision. We additionally present on articulated serial chain manipulators to demonstrate usage on deeper kinematic tree structures. Further media and information can be found at the project website here: \url{https://progress.eecs.umich.edu/projects/splatart/}

\end{abstract}

\section{Introduction}

Robots operating in human centric environments must interact with articulated objects such as drawers, doors, or hand tools. These articulated objects pose significant challenges for robots due to their complex degrees of freedom compared to rigid-body objects, complicating tasks like pose estimation and grasp synthesis. The relative scarcity of suitable datasets exacerbates these challenges. Tools such as LabelFusion \cite{marion2018label} and ProgressLabeller \cite{chen2022progresslabeller} dramatically improve the speed at which researchers can create segmentation or pose estimation datasets (including difficult categories such as transparent objects), but articulated objects remain difficult to scale in terms of data collection. Each additional degree of freedom complicates the representation, and increases the number of observations needed to achieve robust generalization for many tasks. Data driven robot learning approaches would benefit from the ability to generate large numbers of synthetic articulated object observations from a comparatively small number of real-world collected data. Towards that end, representations such as Neural Radiance Fields (NeRFs) \cite{mildenhall2021nerf} and Gaussian Splats \cite{kerbl20233d} have seen success as implicit and explict representations respectively for articulated objects in robotic applications. For example, NARF-22 \cite{lewis2022narf22} showed success in applying articulated NeRFs to a real-world tool pose and configuration estimation task, and GraspSplats \cite{jigraspsplats} demonstrated success in using gaussian splatting for grasp synthesis.

\begin{figure}
    \centering
    \includegraphics[width=\columnwidth]{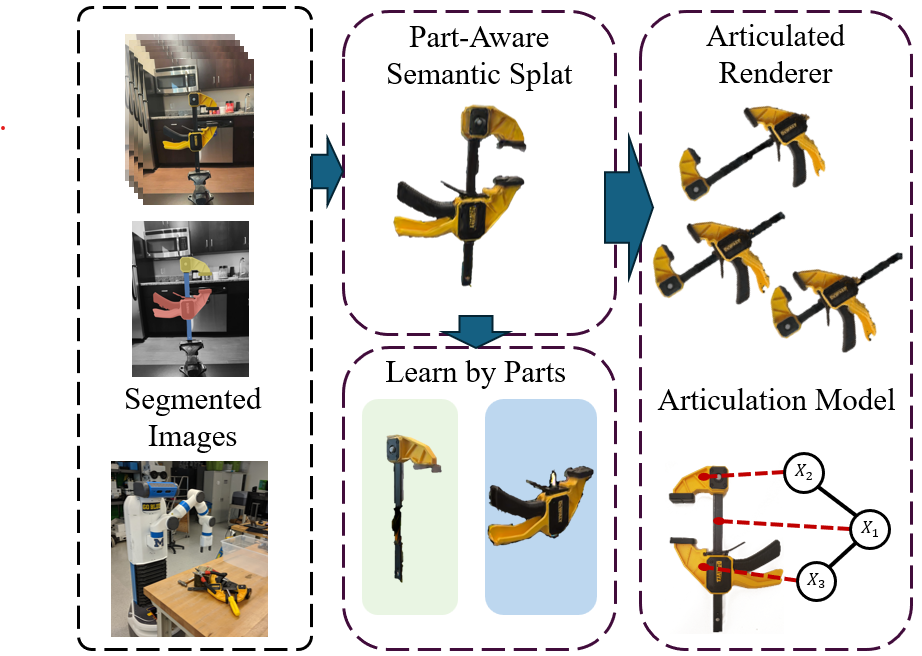}
    \label{fig:pitch_fig}
    \caption{SPLATART is a pipeline which produces articulated gaussian splat representations. The pipeline utilizes cross-scene gaussian splat rendering to estimate part poses and joint parameters from input part-segmented images of an articulated object. From these part poses, Splatart estimates the joint connectivity and motion parameters even for objects with deeper kinematic structures such as robot arms.}
\end{figure}

\begin{figure*}[t]
    \centering
    \includegraphics[width=0.8\linewidth]{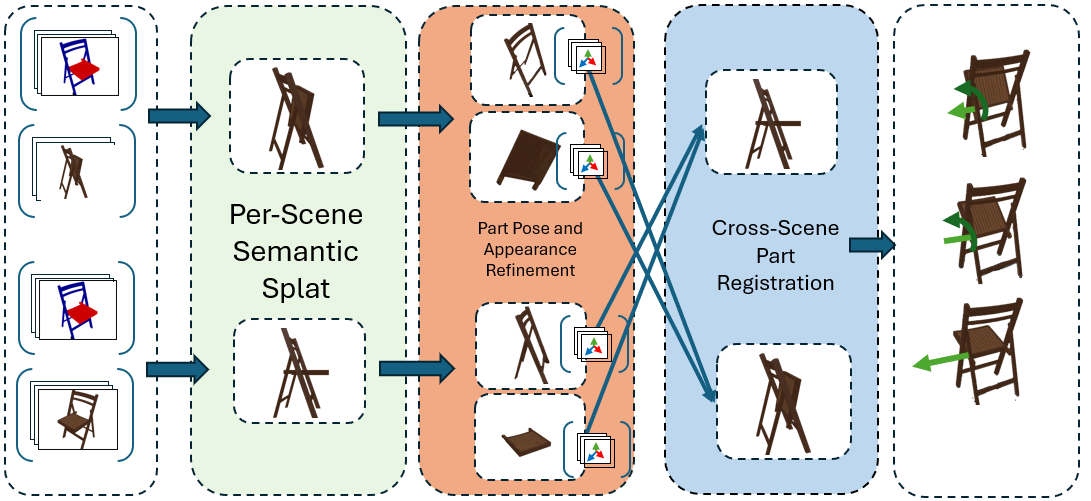}
    \caption{An Overview of the SPLATART Pipeline which takes as input posed images and segmentations of articulated objects. In the first stage, a traditional Gaussian splat is fit to the RGB image information, and then to the semantic segmentation labels. Using these labels, the parts from each scene are first translated into the estimated current part frame, then transformed into the estimated part frame for each other observed scene and evaluated against a rendering loss. This process obtains an accurate pose of each part in the base scenes and improves the geometry representation due to areas that may have been occluded. After part poses have been successfully obtained, the articulation model can be estimated by examining the spatial relations between parts across each scene.}
    \label{fig:pipeline}
\end{figure*}

This work focuses on tree-structured articulated objects with greater kinematic depth than previously examined, in particularly objects such as robot arms which may have six or more joints. In this work, we present SPLATART, a parts based approach to gaussian splatting that learns a configuration conditioned gaussian splat-based renderer for tree-structured articulated objects. SPLATART jointly learns a Gaussian splat representation alongside the part poses in or to estimate object articulation structure from a set of training images of the object at different configurations. At test time, our method can render the object in question at user specified configurations, even outside of those shown during training. We show rendering and structure estimations results on the SAPIEN\cite{xiang2020sapien} objects within the Paris dataset  as compared to both Paris\cite{liu2023paris} and DTA \cite{weng2024neural}. We additionally show exmaples of SPLATART's functionality on more complex articulated objects in the form of simulated robot URDF renderings from a publicly available URDF dataset \cite{URDFDataset}.


\section {Related Work}

Articulated objects remain a difficult class of objects for robots to perceive and manipulate. Earlier works have utilized fiducial markers or similar keypoint trackers to model object articulations through gaussian processes\cite{sturm2011probabilistic} or manifold learning \cite{cohn2022topologically}. Recently, explicit methods using a parts-based approach combining 3d mesh models and URDF (Unified Robotics Description Format) files have seen success such as Pavlasek et al. which leveraged a belief propagation method combined with a a learned-likelihood particle filter to perform pose and configuration estimation on cluttered tool scenes\cite{pavlasek2020parts}. However, creating these models is laborious and difficult. Neural Radiance Fields (NeRF)\cite{mildenhall2021nerf} and Gaussian Splats \cite{kerbl20233d} have showed success in breaking the reliance on mesh models via works such as in NARF22\cite{lewis2022narf22}, NARF24\cite{lewis2024narf24} and Cla-nerf \cite{tseng2022cla} by learning per-part renderers, or by aiming for category level generalization. While these works have continued the parts-based approach to produce NeRF representations of articulated objects, they may still require a-priori knowledge of the articulated object's structure, segmentation, or other difficult to obtain labels. Works such as PARIS \cite{liu2023paris}, DTA \cite{weng2024neural}, and Real2Code \cite{mandi2024real2code} have taking steps at addressing these, but so far they have primarily shown results on objects with smaller kinematic tree depth.

Other works have utilized a more data-driven, deeply learned approaches to infer object articulations. URDFormer \cite{chen2023urdformer} utilized simulation assets combined with known URDF models to learn an image-to-URDF transformer model. Weng el al. additionally inferred joint locations and angles from posed RGB input images via part segmentations \cite{weng2024neural}. Several works have also perceived articulations from video such as GART \cite{Lei_2024_CVPR} which utilized template models to infer articulations even for deformable articulated objects without clear part separations (e.g. a dog or human body). Recent works such as ArtGS \cite{splattingbuilding} have extended gaussian splatting to be more amenable to articulated object representations as well.


\section{Method}



The SPLATART pipeline comprises five primary steps:
\begin{enumerate}
    \item Per-scene semantic gaussian splat training
    \item Cross-scene part pose estimation
    \item Joint parameter estimation
    \item Tree-structure generation
    \item Novel pose/configuration rendering
\end{enumerate}

An overview of this pipeline is shown in Figure \ref{fig:pipeline}, and each component step is discussed in detail below.

\subsection{Problem Formulation}
The problem SPLATART addresses is as follows:
Given $N_v$ posed RGB images of an unknown articulated object and their associated image-space part seperations $\{{\Bar{X}_i^t, \Bar{I}_i^t}, \Bar{S}_i^t\}_{i=1}^{N_v}$ (where $\Bar{X}_i^t$ is image $i,t$'s camera pose, $\Bar{I}_i^t$ is the RGB image data, and $\Bar{S}_i^t$ is the segmentation data) at observation times $t \in \{0,1\}$, SPLATART aims to estimate the per-scene part poses $\{P^t_j\}_{j=1}^{N_p}$ (where $N_p$ is the number of rigid-body component parts of the articulated object), the set of Gaussian parameters $\mathcal{G}^c_j$ that canonically represent each part, and the joint parameters $\mathcal{M}$ that describe the inter-part relationships (e.g. joint axis, rotation angle, etc.). Although SPLATART does not preclude the use of more than two observations (i.e. $t \in \{0,1,...n\}$), this work assumes only two scenes are present.

\subsection{Semantic Gaussian Splat Training}

Given a camera pose Gaussian Splats render an RGB image by rasterizing a collection of Gaussian functions $\mathcal{G} = \{\mu_i,\Sigma_i,\alpha_i,{sh}_i\}$ where $\mu$, $\Sigma$, $\alpha$, and $sh$ represent the means, covariances, opacities and spherical harmonic coefficients of each distribution, and $i$ is an index with $i \in [0,N_{\mathcal{G}}]$ where $N_\mathcal{G}$ is the number of gaussian entries in the splat. In practice, splats typically do not store and optimize $\Sigma_i$, but instead store a collection of scale variables $s_i \in \mathbb{R}^3$ and quaternions $q_i$ which are used to compute the covariance $\Sigma_i = R_is_is_i^TR_i^T$ where $R_i$ is the rotation matrix corresponding to quaternion $q_i$. SPLATART extends this representation to render semantic information by including $\mathcal{S}$, where each $\mathcal{S}_i$ is a vector of length $N_p$. Each index $n \in [0, N]$ of $\mathcal{S}_i$ then corresponds to the likelihood score of a splat belonging to the representation of part $n$'.

SPLATART uses the Splatfacto model as provided in NerfStudio \cite{tancik2023nerfstudio} to train $t$ splats $\mathcal{G}_{t\in \{0,1\}}$, which we will ultimately use to determine the canonical part splats $\{\mathcal{G}^c_j\}$. The Splatfacto model is convenient as it represents a typical collection of Gaussian Splat training improvements such as scale regularization as proposed in PhysGaussians\cite{xie2024physgaussian}. Details of the specifc parameters used are provided in the supplement. The semantic splat representations are trained in a two-step process. First, a Splatfacto model is trained with all gaussian parameters except for $\mathcal{S}$ included as optimization parameters. Subsequently, a secondary optimization is performed where the $\mathcal{S}$ parameters are learned with a traditional Softmax and CrossEntropy loss, but the remaining gaussian parameters are held fixed. This was selected in an attempt to ease the creation of segmentation labels, which are the most laborious data input to generate for SPLATART. Given a subset of the full segmentation mask dataset $\Bar{s}^t \subset \Bar{S}^t$, so long as the total geometry of the articulated object is observed in the labels, the semantic information can still be imposed on the final learned representation, and subsequently used in the remainder of the pipeline.

\subsection{Part Pose Estimation}
Once the segmented gaussian splat has been trained, the next step is to infer the pose of each component part: $\{P^t_j\}_{j=1}^{N_p}$. To do this, a pose is created for every part and scene, paramaterized by a three degree of freedom translation, and a three degree of freedom set of Euler angles: $P^t_j = [T \in \mathbb{R}^3, \theta \in \mathbb{R}^3]$. The translation component for each part is initialized to the Euclidean centroid of the part means, whereas the rotation component can either be initialized to zero, or it can be seeded via an ICP solution between the part means at $t=0$ and the means at $t=1$ to help with numerical stability at the beginning of optimization. 

The part pose parameters, along with the mean, quaternion, scale, and color Gaussian splat parameters $(\mu_i, q_i, s_i, sh_i)$ are iteratively improved via back-propagation and an Adam optimizer \cite{kingma2014adam}. During each iteration, the current estimated transform between each part at $t=0$ and $t=1$ is computed as $(\hat{P}^0_t)^{-1}\hat{P}^1_t$ and $(\hat{P}^1_t)^{-1}\hat{P}^0_t$. Using these estimated transforms, every part at each time observation is moved to the current estimate of the part's pose in the opposite observation. The composition of the parts is rendered as an RGB image, as an accumulation map, and as a segmentation map at the same poses used to train the original splat. These are then used to optimize against the loss functions shown in equation \ref{eq:loss_fn}, where $\hat{RGB}$, $\hat{Acc}$, $\hat{Seg}$ are the predicted accumulation, RGB, and segmentation images. $Acc_{gt}$, $RGB_{gt}$, and $Seg_{gt}$ are their corresponding ground truth images, SSIM denotes the Structural Similarity Index Measure as per Wang et al. \cite{wang2004image}, and $CE$ denotes the standard cross entropy function. The various $\lambda$ hyperparameters are tuned by hand, and represent the tradeoffs between accumulation, RGB, segmentation, and Structural Similarit

\begin{gather}
  \text{loss}_{acc}(\hat{Acc},Acc_{gt}) =  ||\hat{Acc} - Acc_{gt}||_1\\ 
  \text{loss}_{l1}(\hat{RGB},RGB_{gt}) = ||\hat{RGB} - RGB_{gt}||_1\\
  \text{loss}_{ss}(\hat{RGB},RGB_{gt})=1-SSIM(\hat{RGB},RGB_{gt})\\ \text{loss}_{seg}(Seg_{pd}, Seg_{gt}) = CE(Seg_{pd}, Seg_{gt})
\end{gather}
\begin{gather}
  \text{loss}_{total} = \begin{bNiceMatrix}
  \lambda_{acc} & \lambda_{l1} & \lambda_{ssim} & \lambda_{seg}
  \end{bNiceMatrix}
  \begin{bNiceMatrix}
      \text{loss}_{acc}\\
      \text{loss}_{rgb}\\
      \text{loss}_{ssim}\\
      \text{loss}_{seg}
  \end{bNiceMatrix}
  \label{eq:loss_fn}
\end{gather}

The inclusion of both the Gaussian and pose parameters in the optimization routine is important for two primary reasons. Firstly, it is the primary mechanism by which parts are able to model geometry that was not visible in their original scene. In order for the loss to improve, each part's Gaussian splat representaiton must evolve so that the part can produce quality renderings in every observed configuration of the object. Secondly, gradient descent on pose is prone to stalling in local minima. We heuristically observed that including the Gaussian parameters during optimization improved convergence speed and final answer quality by providing a straighter path to the solution through the parameter space, similar to the pattern observed by Lin et al. \cite{lin2023parallel}.

\begin{figure*}[t]
    \begin{subfigure}[t]{0.1\linewidth}
    \includegraphics[width=\linewidth, height=\textheight, keepaspectratio]{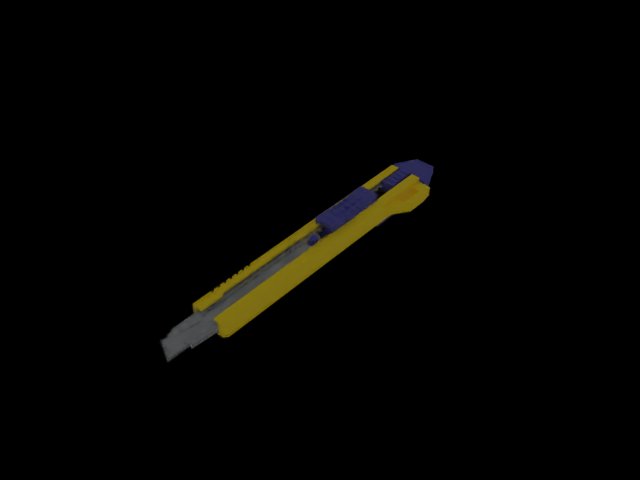}
    \end{subfigure}
    \begin{subfigure}[t]{0.1\linewidth}
    \includegraphics[width=\linewidth, height=\textheight, keepaspectratio]{media/blade/rendered_image_0.png}
    \end{subfigure}
    \begin{subfigure}[t]{0.1\linewidth}
    \includegraphics[width=\linewidth, height=\textheight, keepaspectratio]{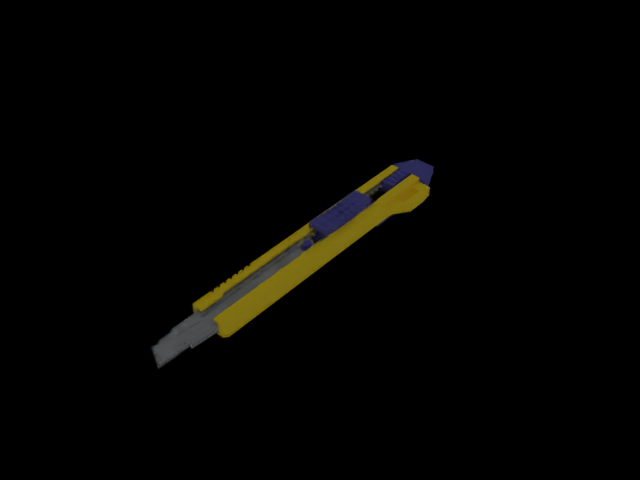}
    \end{subfigure}
    \begin{subfigure}[t]{0.1\linewidth}
    \includegraphics[width=\linewidth, height=\textheight, keepaspectratio]{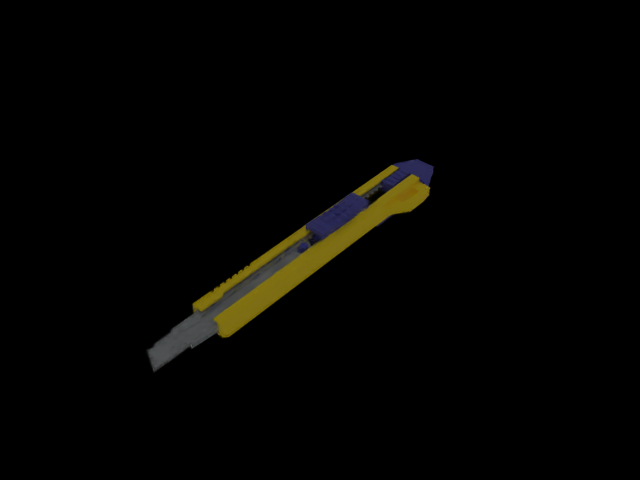}
    \end{subfigure}
    \begin{subfigure}[t]{0.1\linewidth}
    \includegraphics[width=\linewidth, height=\textheight, keepaspectratio]{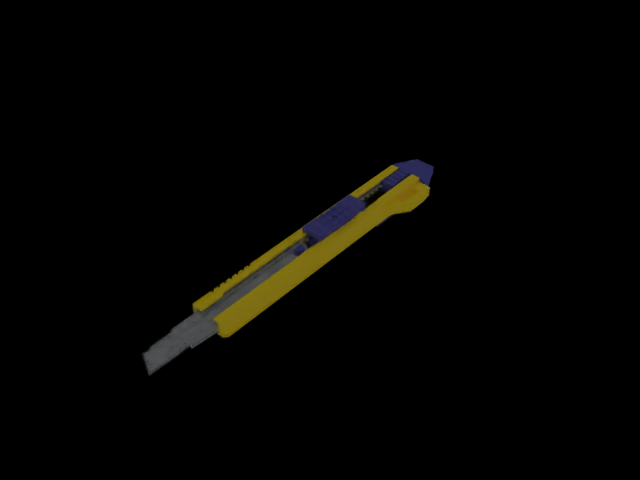}
    \end{subfigure}
    \begin{subfigure}[t]{0.1\linewidth}
    \includegraphics[width=\linewidth, height=\textheight, keepaspectratio]{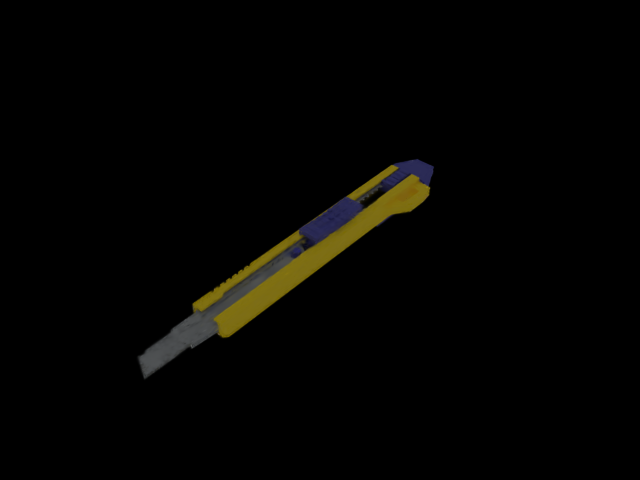}
    \end{subfigure}
    \begin{subfigure}[t]{0.1\linewidth}
    \includegraphics[width=\linewidth, height=\textheight, keepaspectratio]{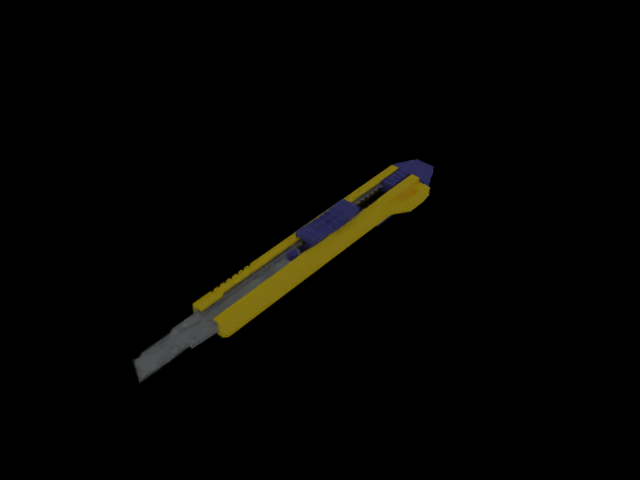}
    \end{subfigure}
    \begin{subfigure}[t]{0.1\linewidth}
    \includegraphics[width=\linewidth, height=\textheight, keepaspectratio]{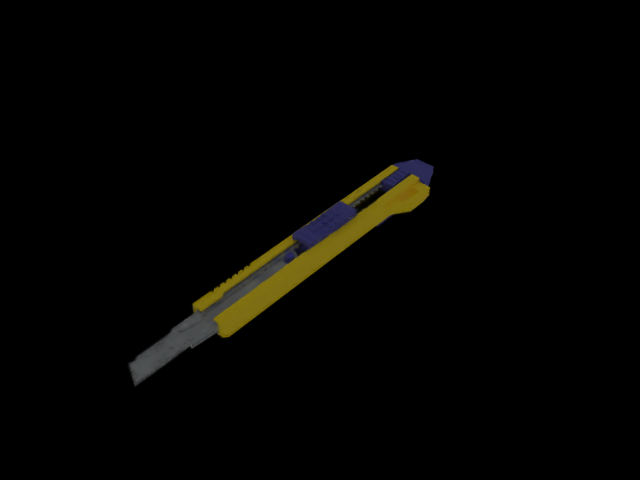}
    \end{subfigure}
    \begin{subfigure}[t]{0.1\linewidth}
    \includegraphics[width=\linewidth, height=\textheight, keepaspectratio]{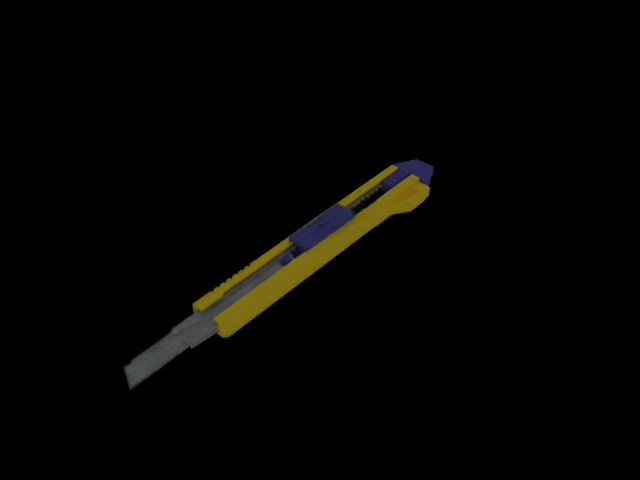}
    \end{subfigure}

    \vspace{0.1cm}
    \begin{subfigure}[t]{0.1\linewidth}
    \includegraphics[width=\linewidth, height=\textheight, keepaspectratio]{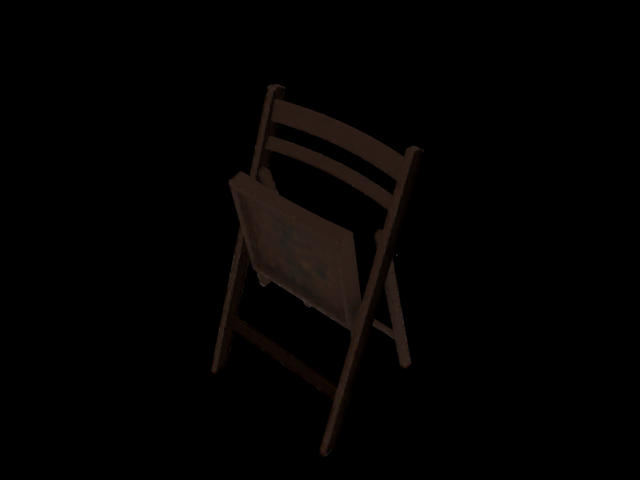}
    \end{subfigure}
    \begin{subfigure}[t]{0.1\linewidth}
    \includegraphics[width=\linewidth, height=\textheight, keepaspectratio]{media/foldchair/rendered_image_0.png}
    \end{subfigure}
    \begin{subfigure}[t]{0.1\linewidth}
    \includegraphics[width=\linewidth, height=\textheight, keepaspectratio]{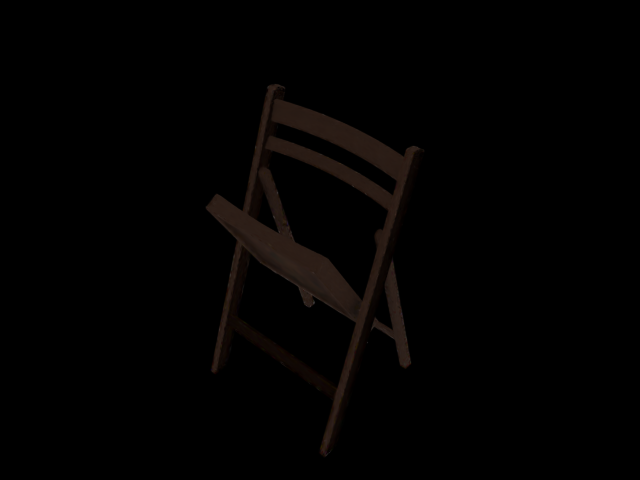}
    \end{subfigure}
    \begin{subfigure}[t]{0.1\linewidth}
    \includegraphics[width=\linewidth, height=\textheight, keepaspectratio]{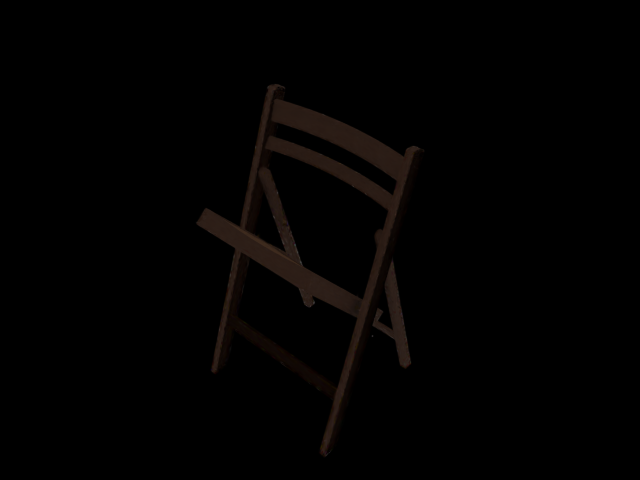}
    \end{subfigure}
    \begin{subfigure}[t]{0.1\linewidth}
    \includegraphics[width=\linewidth, height=\textheight, keepaspectratio]{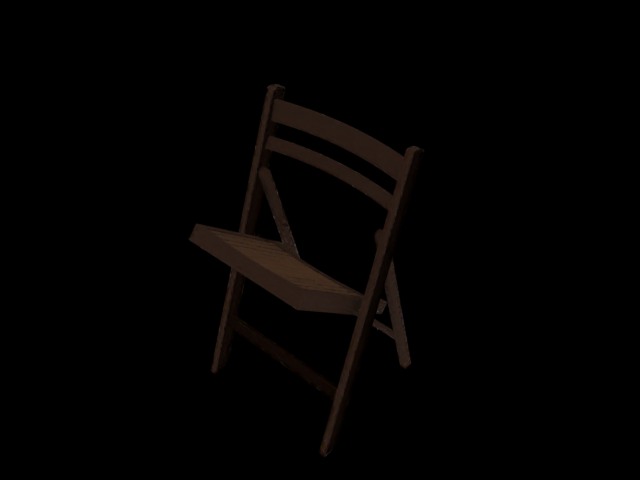}
    \end{subfigure}
    \begin{subfigure}[t]{0.1\linewidth}
    \includegraphics[width=\linewidth, height=\textheight, keepaspectratio]{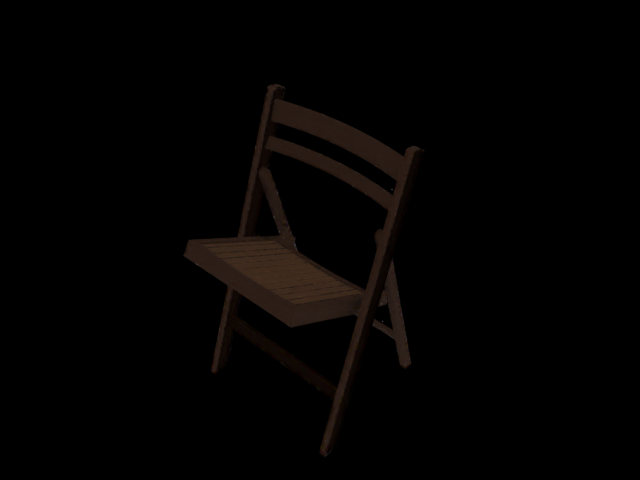}
    \end{subfigure}
    \begin{subfigure}[t]{0.1\linewidth}
    \includegraphics[width=\linewidth, height=\textheight, keepaspectratio]{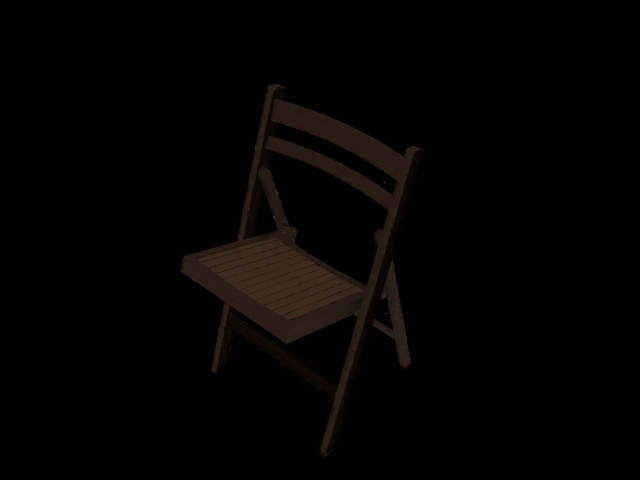}
    \end{subfigure}
    \begin{subfigure}[t]{0.1\linewidth}
    \includegraphics[width=\linewidth, height=\textheight, keepaspectratio]{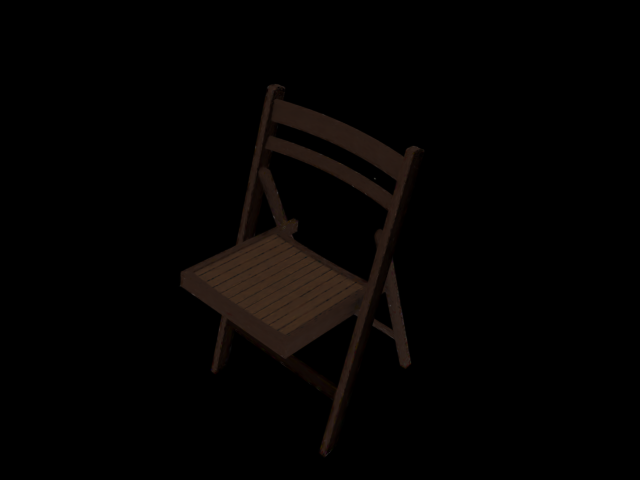}
    \end{subfigure}
    \begin{subfigure}[t]{0.1\linewidth}
    \includegraphics[width=\linewidth, height=\textheight, keepaspectratio]{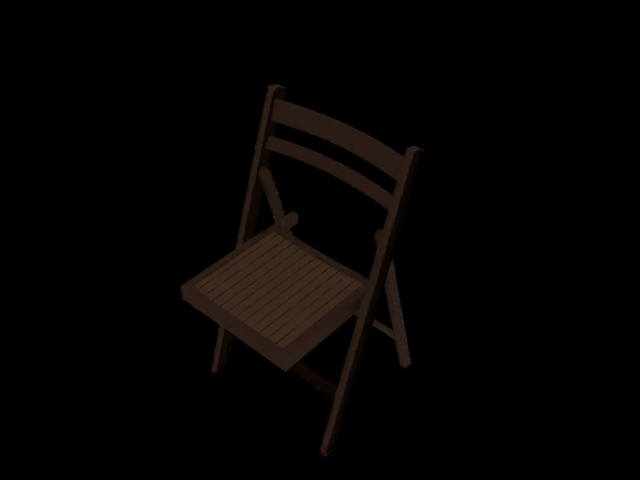}
    \end{subfigure}

    \vspace{0.1cm}
    \begin{subfigure}[t]{0.1\linewidth}
    \includegraphics[width=\linewidth, height=\textheight, keepaspectratio]{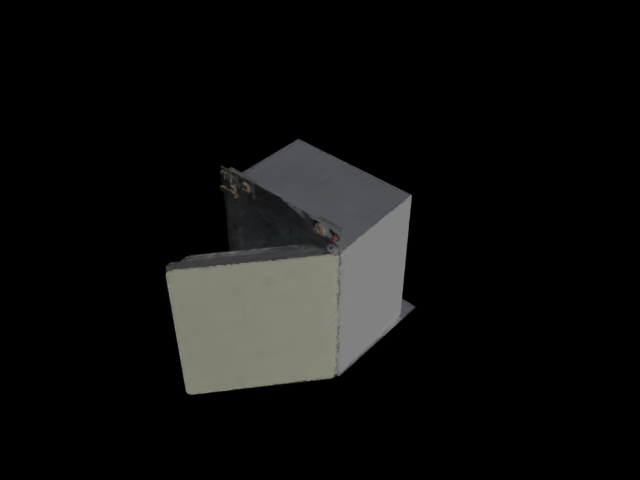}
    \end{subfigure}
    \begin{subfigure}[t]{0.1\linewidth}
    \includegraphics[width=\linewidth, height=\textheight, keepaspectratio]{media/fridge/rendered_image_0.png}
    \end{subfigure}
    \begin{subfigure}[t]{0.1\linewidth}
    \includegraphics[width=\linewidth, height=\textheight, keepaspectratio]{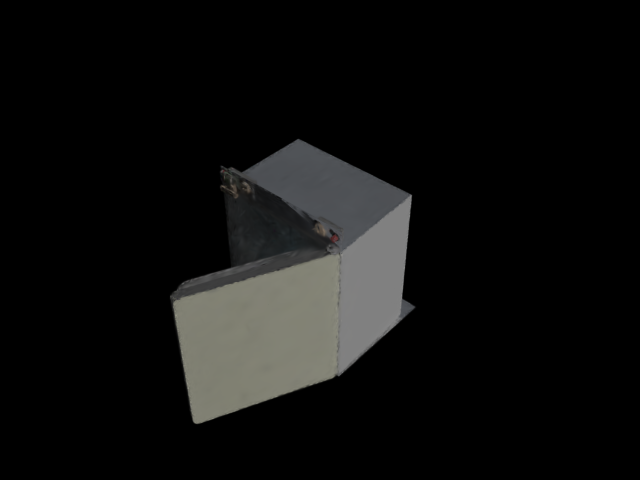}
    \end{subfigure}
    \begin{subfigure}[t]{0.1\linewidth}
    \includegraphics[width=\linewidth, height=\textheight, keepaspectratio]{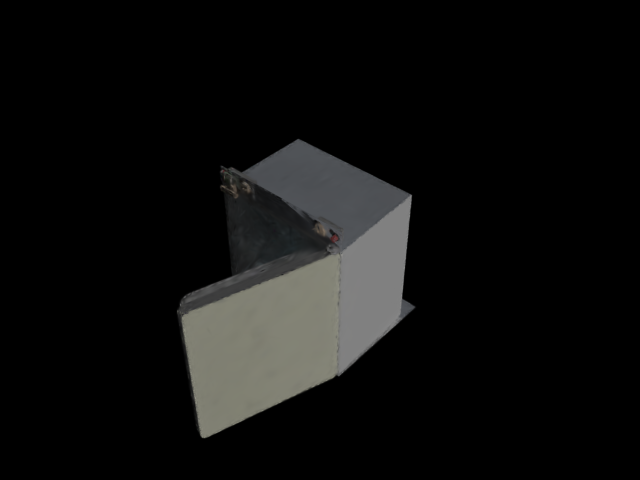}
    \end{subfigure}
    \begin{subfigure}[t]{0.1\linewidth}
    \includegraphics[width=\linewidth, height=\textheight, keepaspectratio]{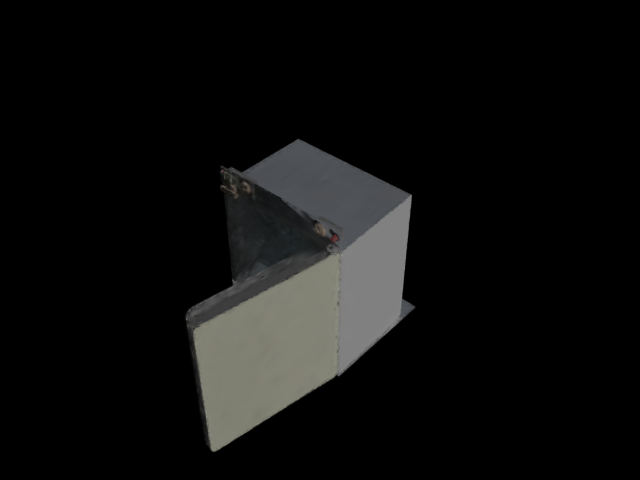}
    \end{subfigure}
    \begin{subfigure}[t]{0.1\linewidth}
    \includegraphics[width=\linewidth, height=\textheight, keepaspectratio]{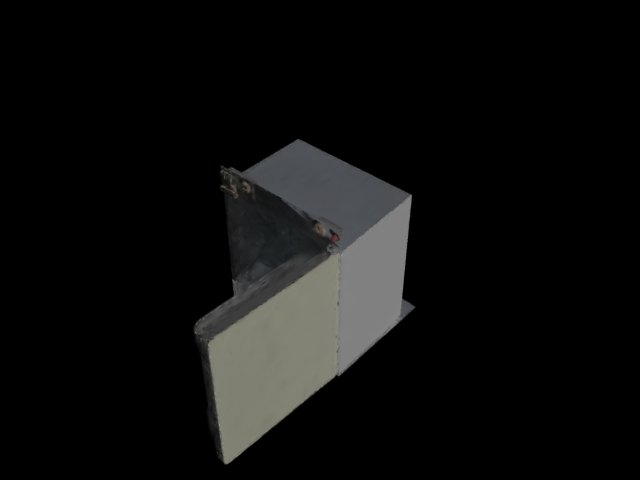}
    \end{subfigure}
    \begin{subfigure}[t]{0.1\linewidth}
    \includegraphics[width=\linewidth, height=\textheight, keepaspectratio]{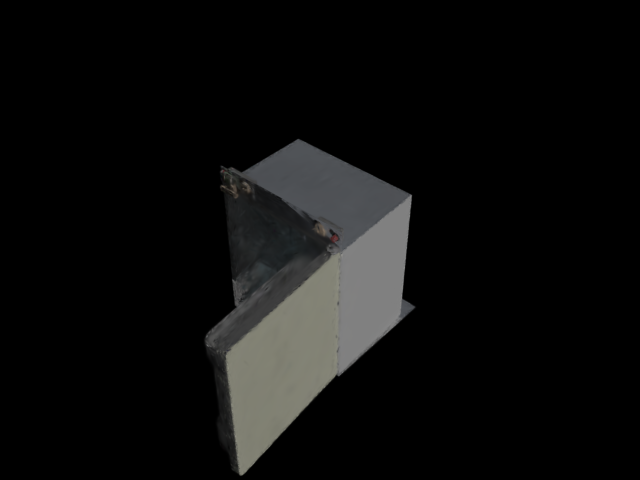}
    \end{subfigure}
    \begin{subfigure}[t]{0.1\linewidth}
    \includegraphics[width=\linewidth, height=\textheight, keepaspectratio]{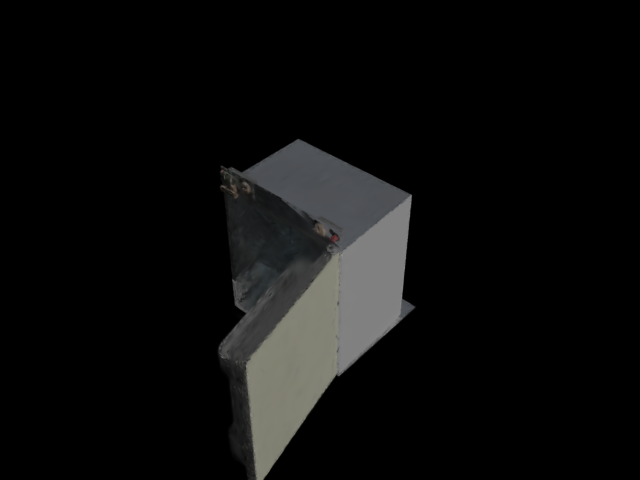}
    \end{subfigure}
    \begin{subfigure}[t]{0.1\linewidth}
    \includegraphics[width=\linewidth, height=\textheight, keepaspectratio]{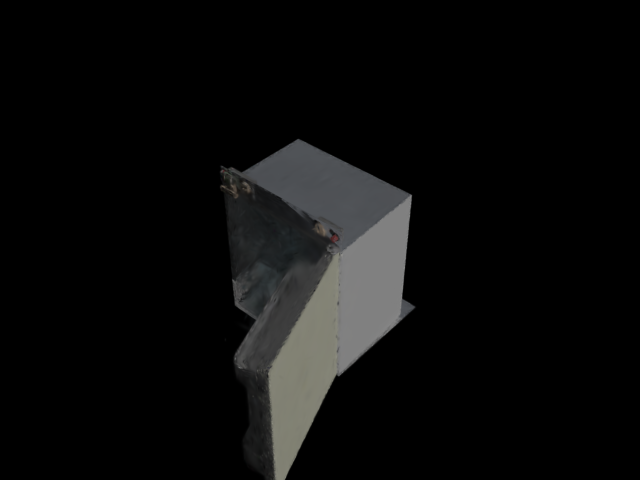}
    \end{subfigure}

    \vspace{0.1cm}
    \begin{subfigure}[t]{0.1\linewidth}
    \includegraphics[width=\linewidth, height=\textheight, keepaspectratio]{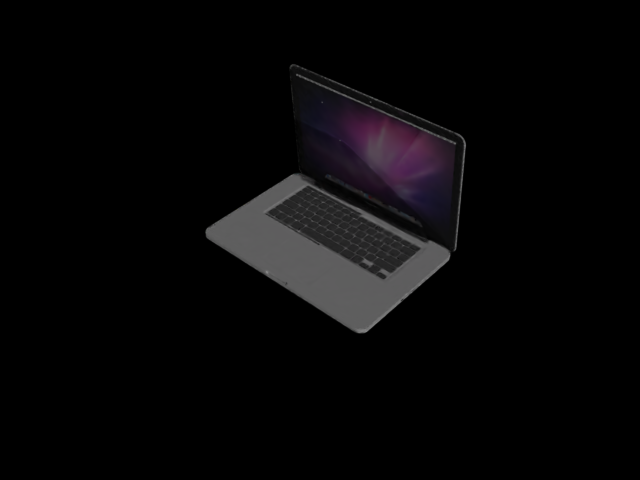}
    \end{subfigure}
    \begin{subfigure}[t]{0.1\linewidth}
    \includegraphics[width=\linewidth, height=\textheight, keepaspectratio]{media/laptop/rendered_image_0.png}
    \end{subfigure}
    \begin{subfigure}[t]{0.1\linewidth}
    \includegraphics[width=\linewidth, height=\textheight, keepaspectratio]{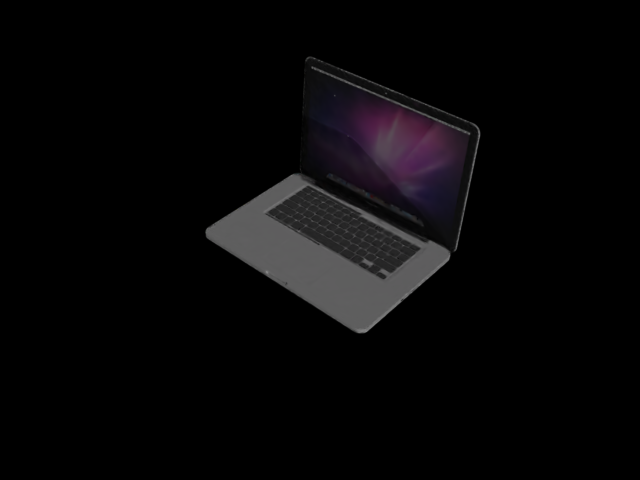}
    \end{subfigure}
    \begin{subfigure}[t]{0.1\linewidth}
    \includegraphics[width=\linewidth, height=\textheight, keepaspectratio]{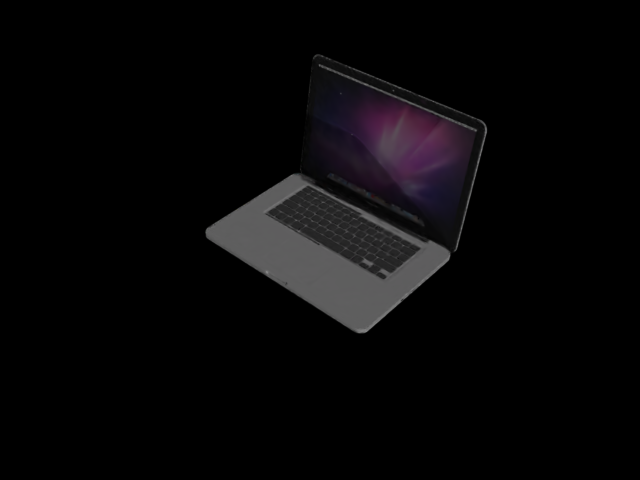}
    \end{subfigure}
    \begin{subfigure}[t]{0.1\linewidth}
    \includegraphics[width=\linewidth, height=\textheight, keepaspectratio]{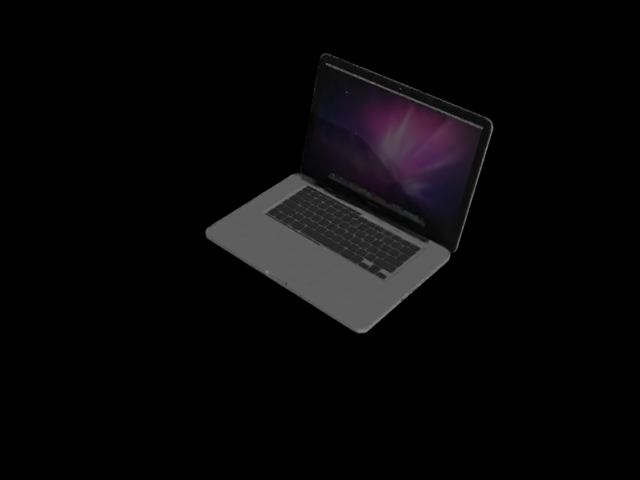}
    \end{subfigure}
    \begin{subfigure}[t]{0.1\linewidth}
    \includegraphics[width=\linewidth, height=\textheight, keepaspectratio]{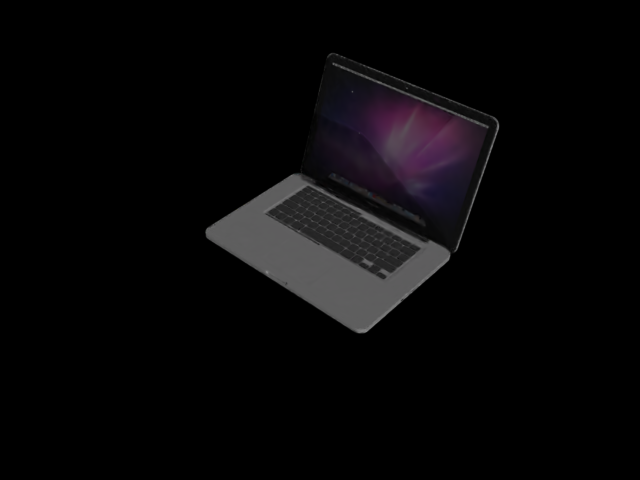}
    \end{subfigure}
    \begin{subfigure}[t]{0.1\linewidth}
    \includegraphics[width=\linewidth, height=\textheight, keepaspectratio]{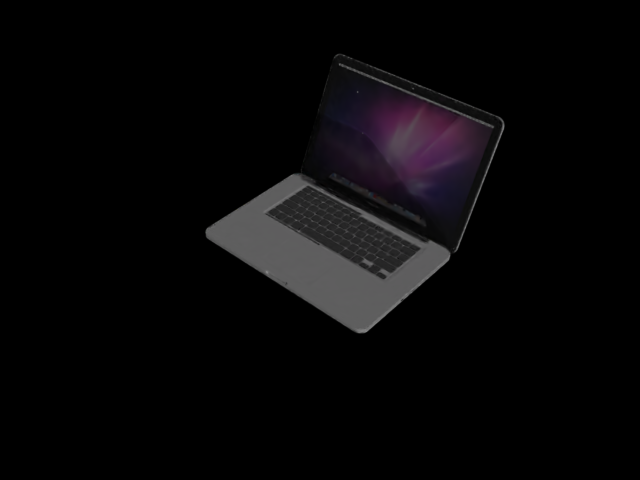}
    \end{subfigure}
    \begin{subfigure}[t]{0.1\linewidth}
    \includegraphics[width=\linewidth, height=\textheight, keepaspectratio]{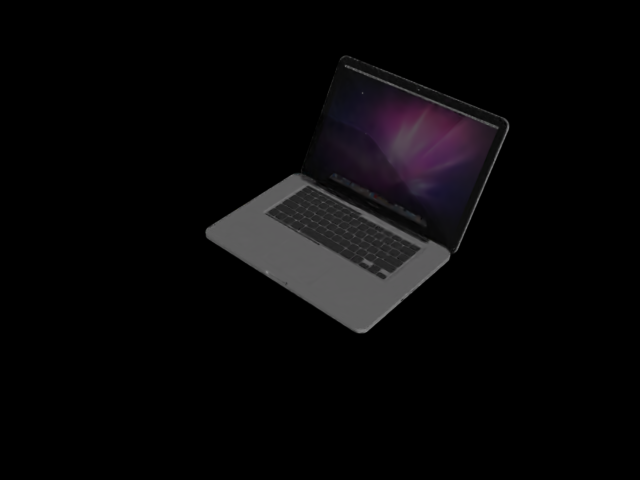}
    \end{subfigure}
    \begin{subfigure}[t]{0.1\linewidth}
    \includegraphics[width=\linewidth, height=\textheight, keepaspectratio]{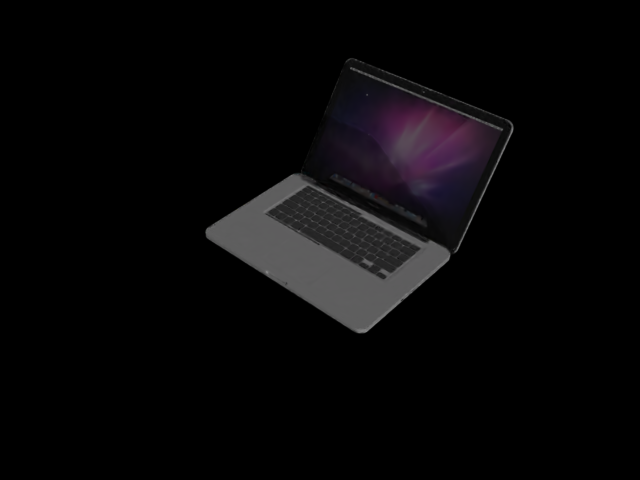}
    \end{subfigure}

    \vspace{0.1cm}
    \begin{subfigure}[t]{0.1\linewidth}
    \includegraphics[width=\linewidth, height=\textheight, keepaspectratio]{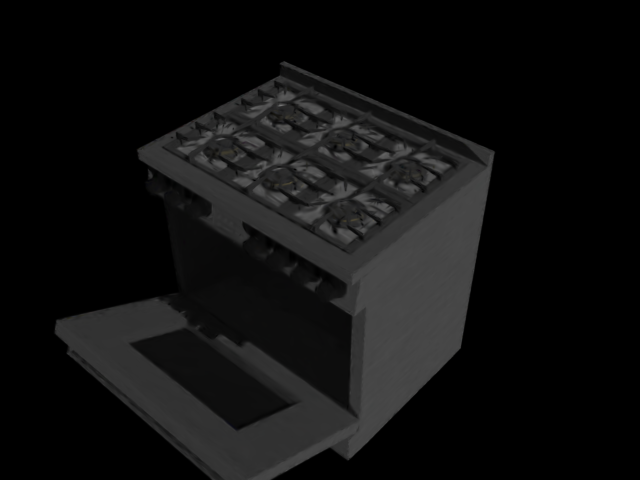}
    \end{subfigure}
    \begin{subfigure}[t]{0.1\linewidth}
    \includegraphics[width=\linewidth, height=\textheight, keepaspectratio]{media/oven/rendered_image_0.png}
    \end{subfigure}
    \begin{subfigure}[t]{0.1\linewidth}
    \includegraphics[width=\linewidth, height=\textheight, keepaspectratio]{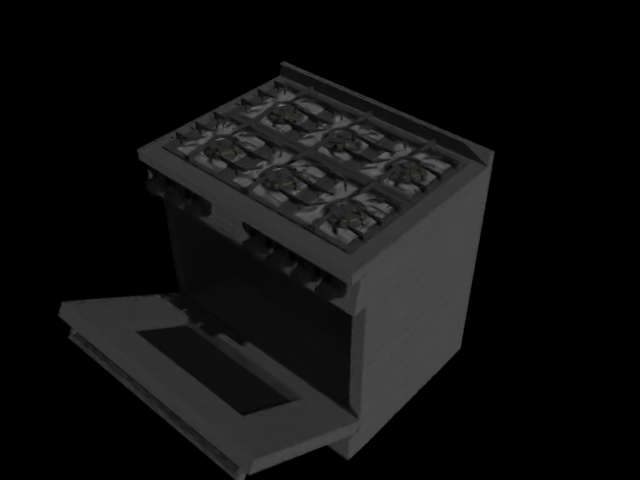}
    \end{subfigure}
    \begin{subfigure}[t]{0.1\linewidth}
    \includegraphics[width=\linewidth, height=\textheight, keepaspectratio]{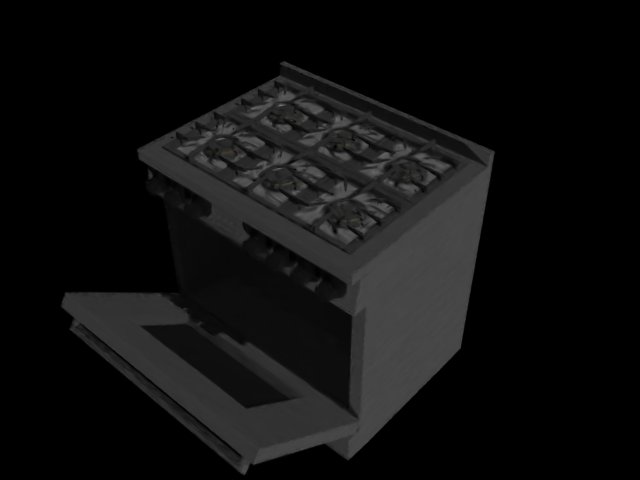}
    \end{subfigure}
    \begin{subfigure}[t]{0.1\linewidth}
    \includegraphics[width=\linewidth, height=\textheight, keepaspectratio]{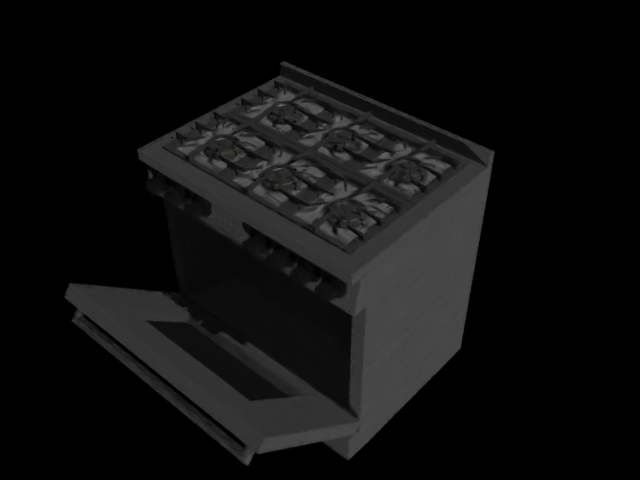}
    \end{subfigure}
    \begin{subfigure}[t]{0.1\linewidth}
    \includegraphics[width=\linewidth, height=\textheight, keepaspectratio]{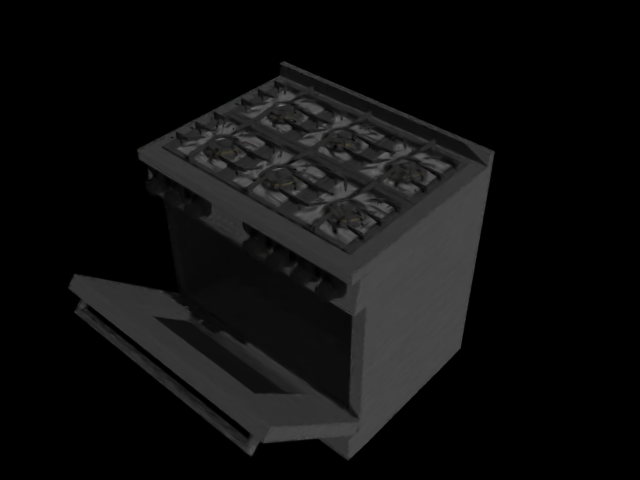}
    \end{subfigure}
    \begin{subfigure}[t]{0.1\linewidth}
    \includegraphics[width=\linewidth, height=\textheight, keepaspectratio]{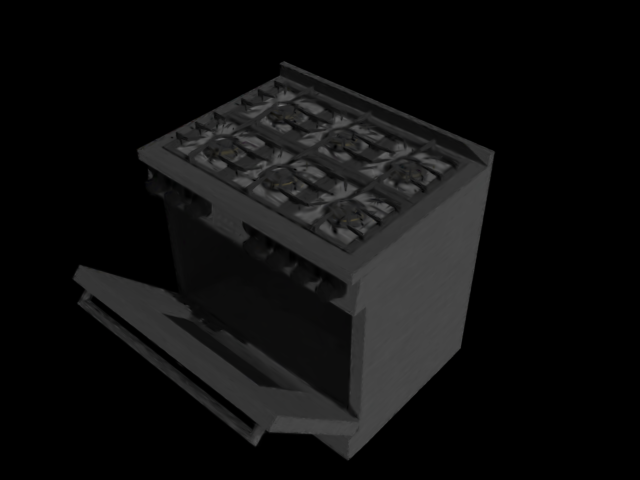}
    \end{subfigure}
    \begin{subfigure}[t]{0.1\linewidth}
    \includegraphics[width=\linewidth, height=\textheight, keepaspectratio]{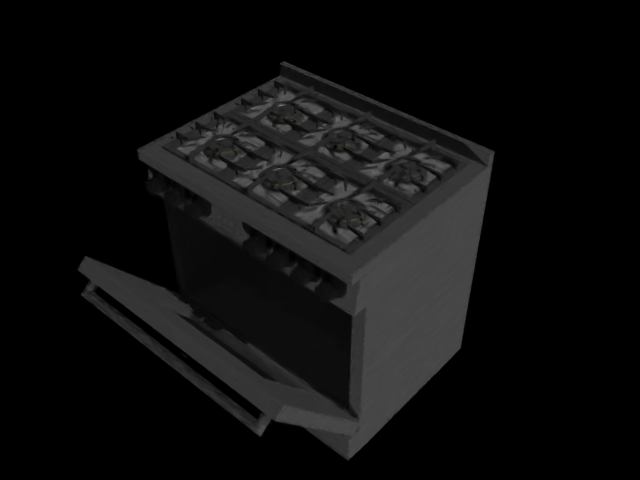}
    \end{subfigure}
    \begin{subfigure}[t]{0.1\linewidth}
    \includegraphics[width=\linewidth, height=\textheight, keepaspectratio]{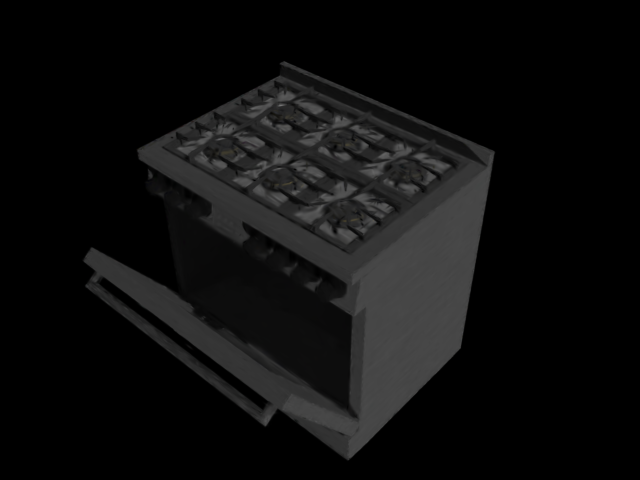}
    \end{subfigure}

    \vspace{0.1cm}
    \begin{subfigure}[t]{0.1\linewidth}
    \includegraphics[width=\linewidth, height=\textheight, keepaspectratio]{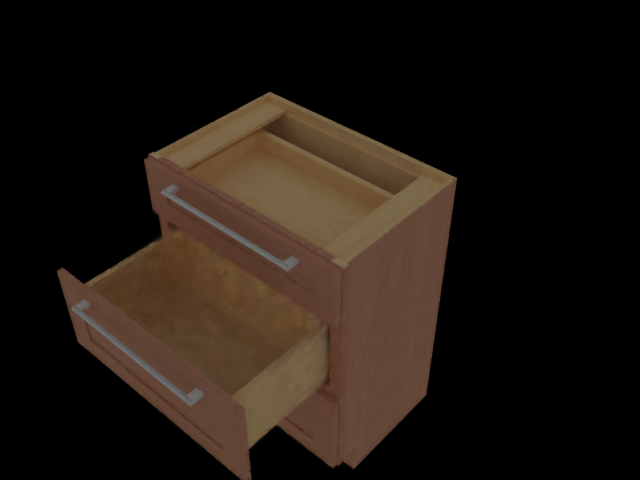}
    \end{subfigure}
    \begin{subfigure}[t]{0.1\linewidth}
    \includegraphics[width=\linewidth, height=\textheight, keepaspectratio]{media/storage/rendered_image_0.png}
    \end{subfigure}
    \begin{subfigure}[t]{0.1\linewidth}
    \includegraphics[width=\linewidth, height=\textheight, keepaspectratio]{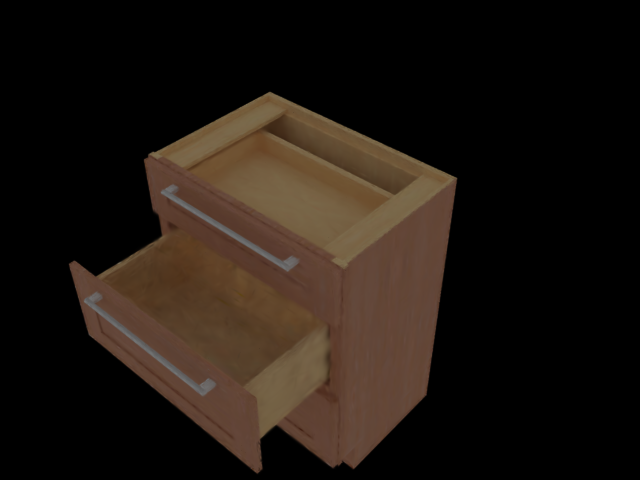}
    \end{subfigure}
    \begin{subfigure}[t]{0.1\linewidth}
    \includegraphics[width=\linewidth, height=\textheight, keepaspectratio]{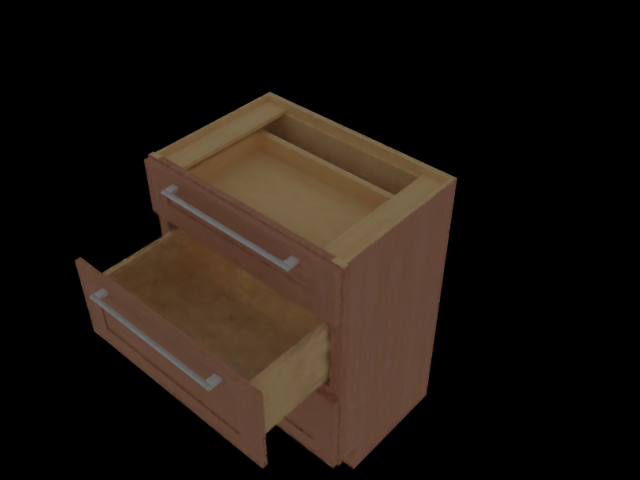}
    \end{subfigure}
    \begin{subfigure}[t]{0.1\linewidth}
    \includegraphics[width=\linewidth, height=\textheight, keepaspectratio]{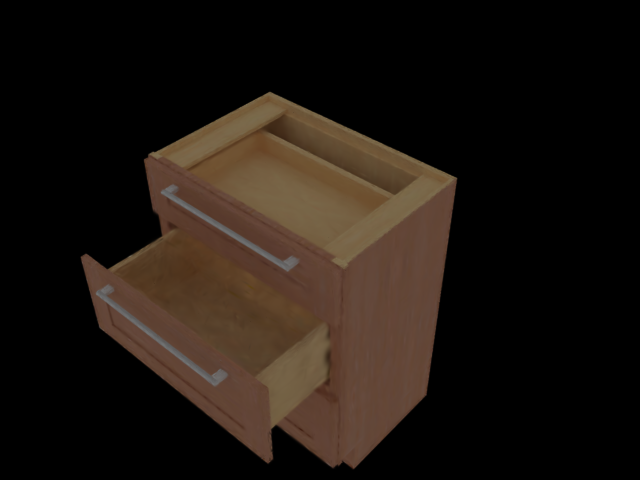}
    \end{subfigure}
    \begin{subfigure}[t]{0.1\linewidth}
    \includegraphics[width=\linewidth, height=\textheight, keepaspectratio]{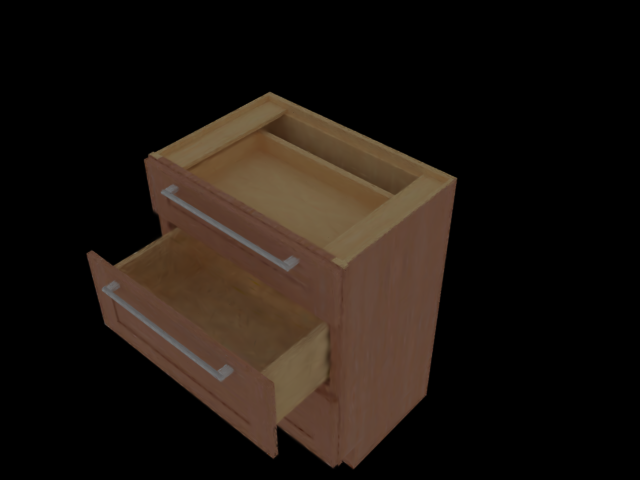}
    \end{subfigure}
    \begin{subfigure}[t]{0.1\linewidth}
    \includegraphics[width=\linewidth, height=\textheight, keepaspectratio]{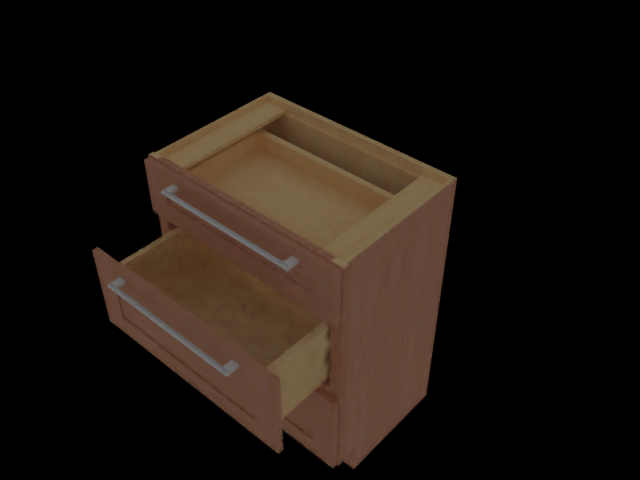}
    \end{subfigure}
    \begin{subfigure}[t]{0.1\linewidth}
    \includegraphics[width=\linewidth, height=\textheight, keepaspectratio]{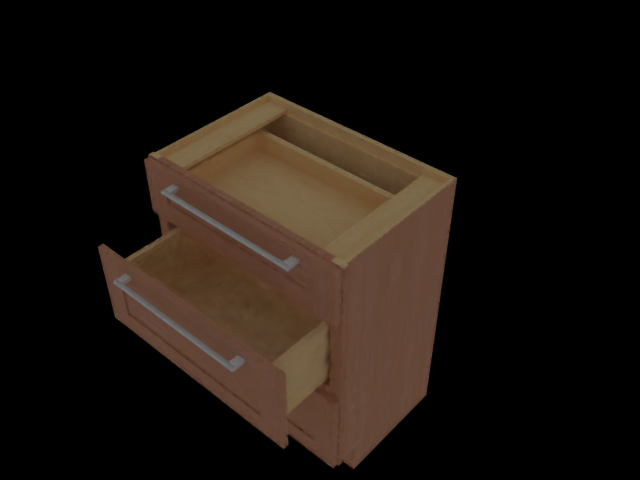}
    \end{subfigure}
    \begin{subfigure}[t]{0.1\linewidth}
    \includegraphics[width=\linewidth, height=\textheight, keepaspectratio]{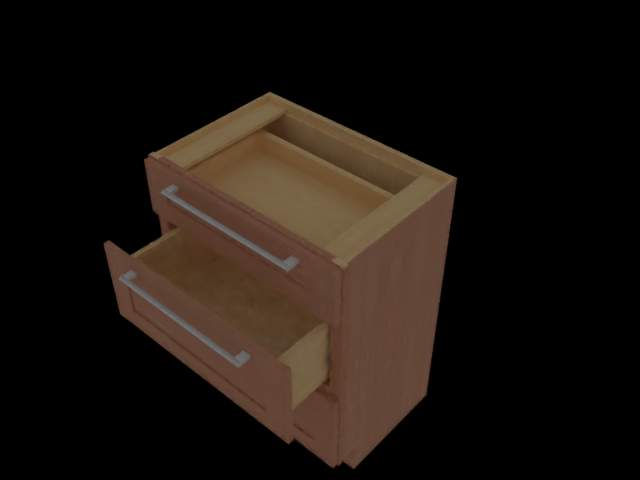}
    \end{subfigure}

    \vspace{0.1cm}
    \begin{subfigure}[t]{0.1\linewidth}
    \includegraphics[width=\linewidth, height=\textheight, keepaspectratio]{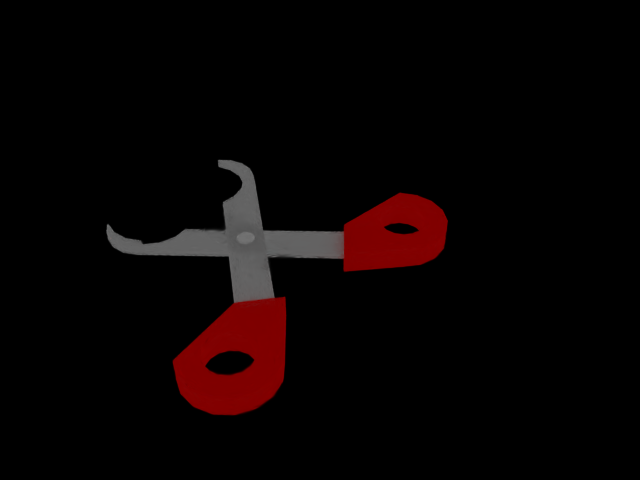}
    \end{subfigure}
    \begin{subfigure}[t]{0.1\linewidth}
    \includegraphics[width=\linewidth, height=\textheight, keepaspectratio]{media/scissor/rendered_image_0.png}
    \end{subfigure}
    \begin{subfigure}[t]{0.1\linewidth}
    \includegraphics[width=\linewidth, height=\textheight, keepaspectratio]{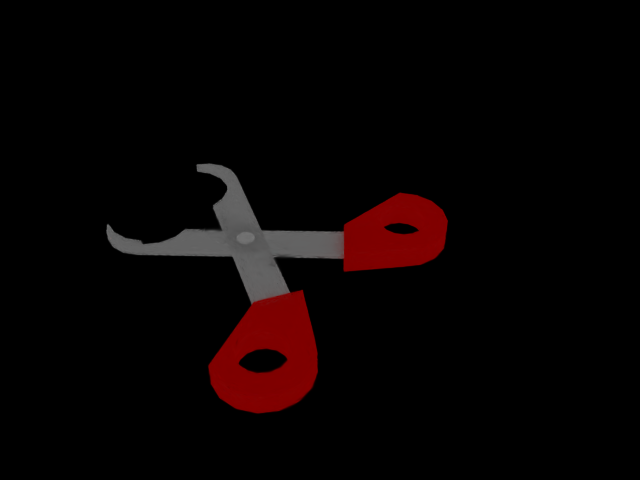}
    \end{subfigure}
    \begin{subfigure}[t]{0.1\linewidth}
    \includegraphics[width=\linewidth, height=\textheight, keepaspectratio]{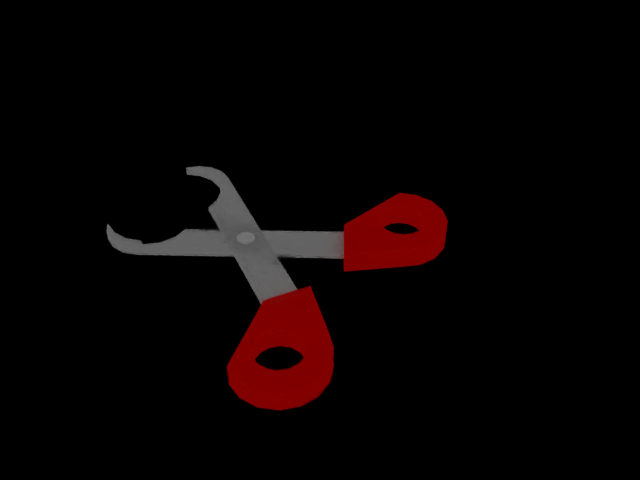}
    \end{subfigure}
    \begin{subfigure}[t]{0.1\linewidth}
    \includegraphics[width=\linewidth, height=\textheight, keepaspectratio]{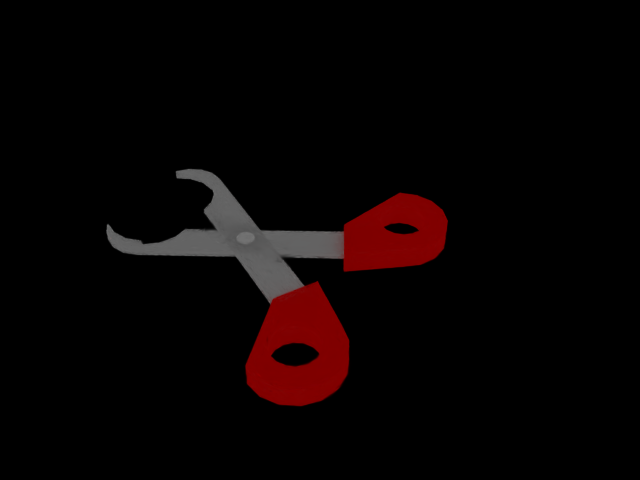}
    \end{subfigure}
    \begin{subfigure}[t]{0.1\linewidth}
    \includegraphics[width=\linewidth, height=\textheight, keepaspectratio]{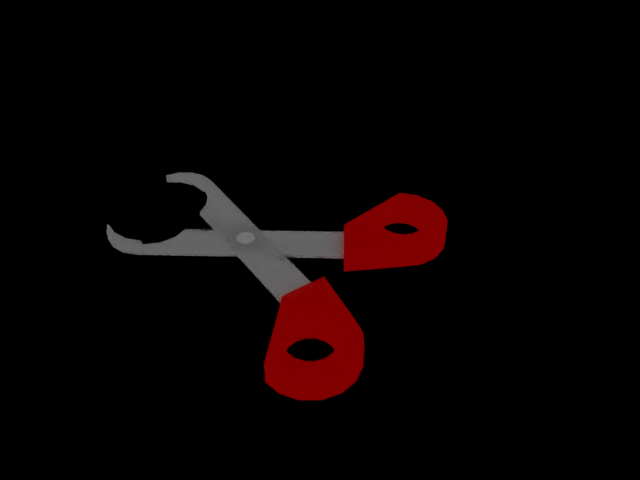}
    \end{subfigure}
    \begin{subfigure}[t]{0.1\linewidth}
    \includegraphics[width=\linewidth, height=\textheight, keepaspectratio]{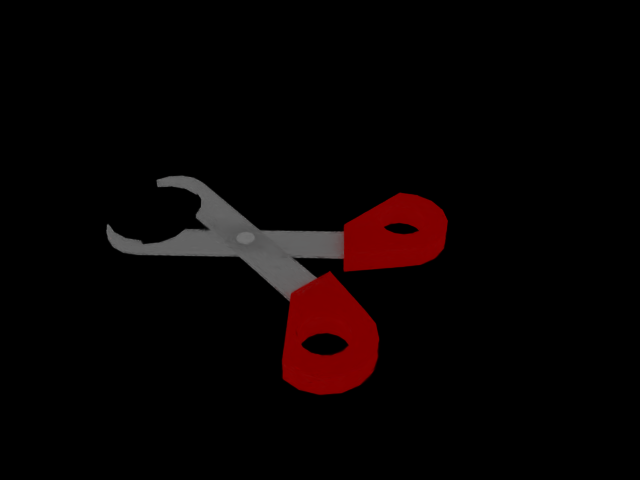}
    \end{subfigure}
    \begin{subfigure}[t]{0.1\linewidth}
    \includegraphics[width=\linewidth, height=\textheight, keepaspectratio]{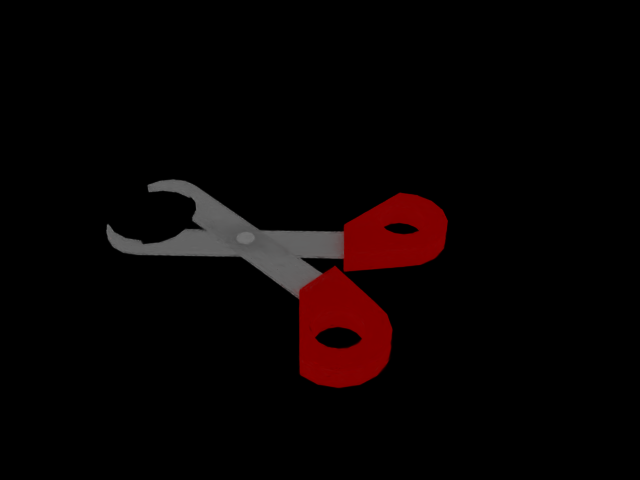}
    \end{subfigure}
    \begin{subfigure}[t]{0.1\linewidth}
    \includegraphics[width=\linewidth, height=\textheight, keepaspectratio]{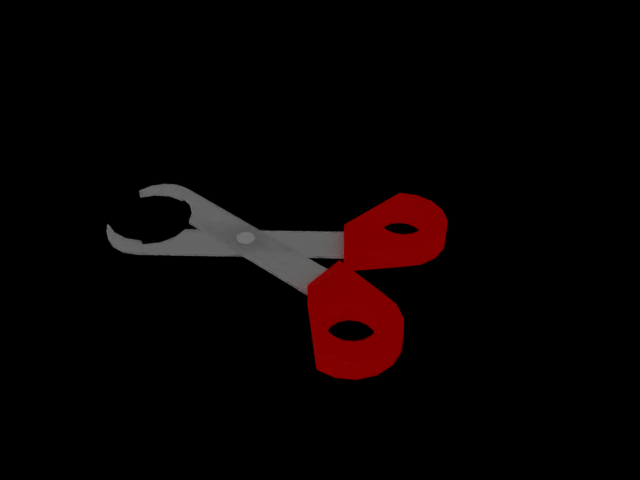}
    \end{subfigure}

    \vspace{0.1cm}
    \begin{subfigure}[t]{0.1\linewidth}
    \includegraphics[width=\linewidth, height=\textheight, keepaspectratio]{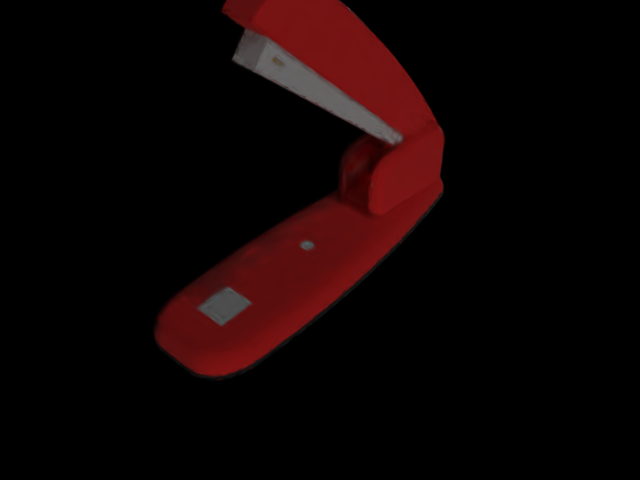}
    \end{subfigure}
    \begin{subfigure}[t]{0.1\linewidth}
    \includegraphics[width=\linewidth, height=\textheight, keepaspectratio]{media/stapler/rendered_image_0.png}
    \end{subfigure}
    \begin{subfigure}[t]{0.1\linewidth}
    \includegraphics[width=\linewidth, height=\textheight, keepaspectratio]{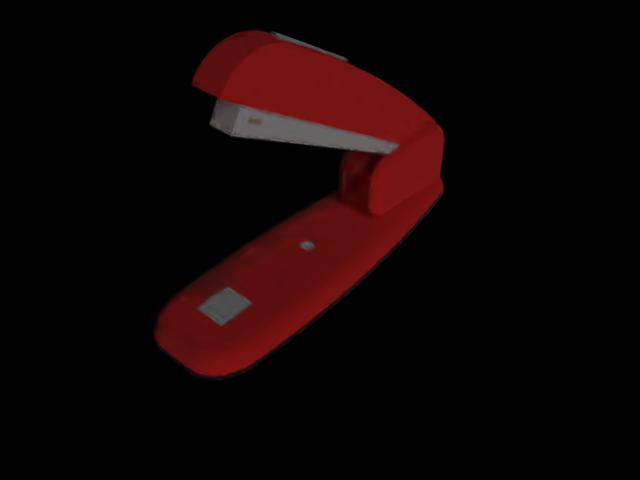}
    \end{subfigure}
    \begin{subfigure}[t]{0.1\linewidth}
    \includegraphics[width=\linewidth, height=\textheight, keepaspectratio]{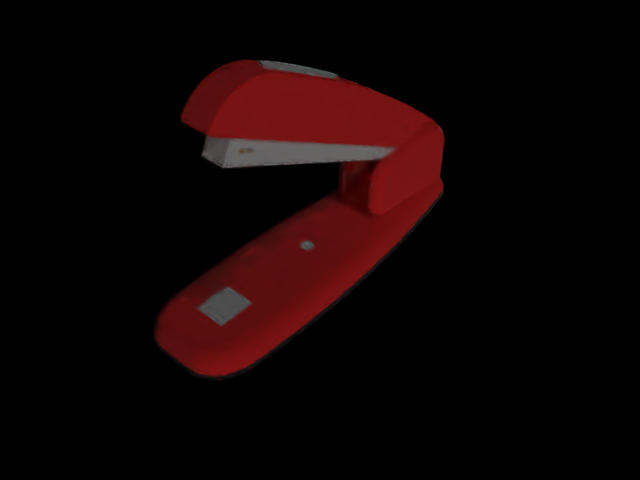}
    \end{subfigure}
    \begin{subfigure}[t]{0.1\linewidth}
    \includegraphics[width=\linewidth, height=\textheight, keepaspectratio]{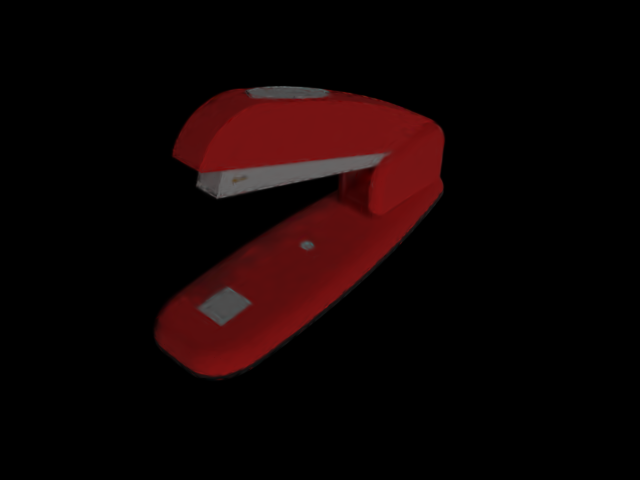}
    \end{subfigure}
    \begin{subfigure}[t]{0.1\linewidth}
    \includegraphics[width=\linewidth, height=\textheight, keepaspectratio]{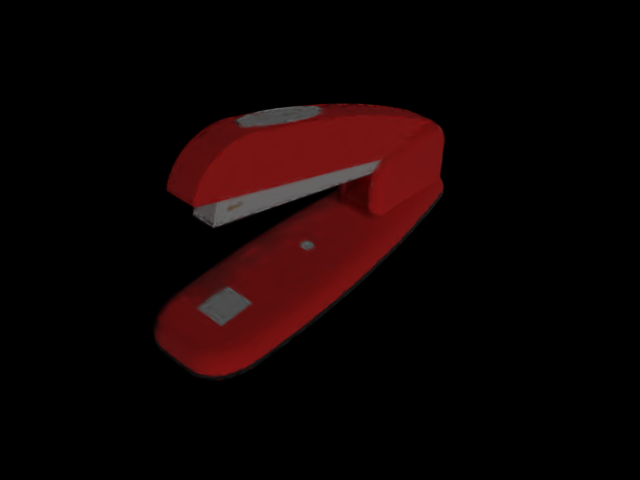}
    \end{subfigure}
    \begin{subfigure}[t]{0.1\linewidth}
    \includegraphics[width=\linewidth, height=\textheight, keepaspectratio]{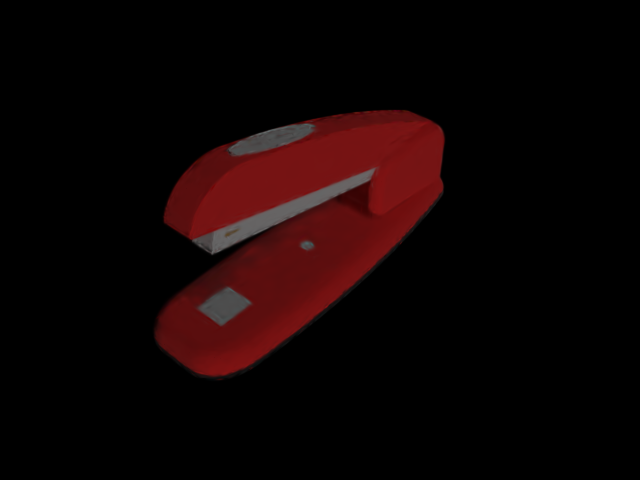}
    \end{subfigure}
    \begin{subfigure}[t]{0.1\linewidth}
    \includegraphics[width=\linewidth, height=\textheight, keepaspectratio]{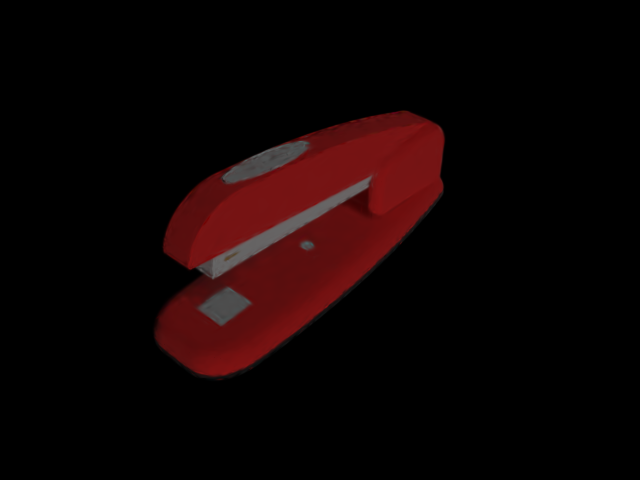}
    \end{subfigure}
    \begin{subfigure}[t]{0.1\linewidth}
    \includegraphics[width=\linewidth, height=\textheight, keepaspectratio]{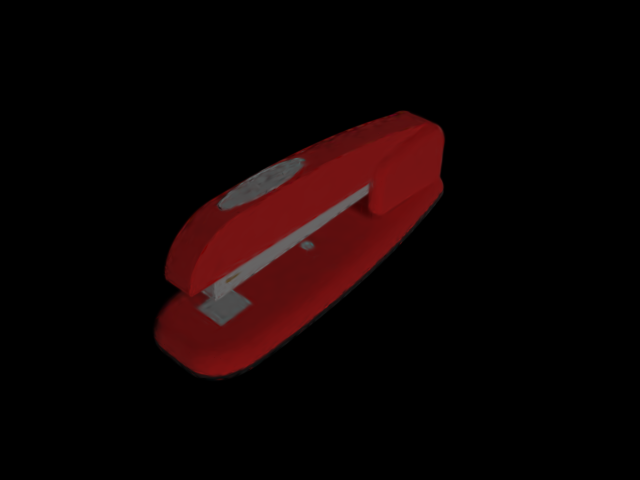}
    \end{subfigure}

    \vspace{0.1cm}
    \begin{subfigure}[t]{0.1\linewidth}
    \includegraphics[width=\linewidth, height=\textheight, keepaspectratio]{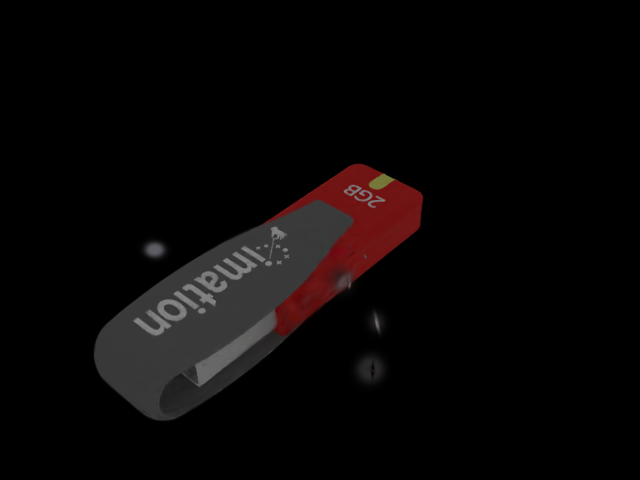}
    \end{subfigure}
    \begin{subfigure}[t]{0.1\linewidth}
    \includegraphics[width=\linewidth, height=\textheight, keepaspectratio]{media/usb/rendered_image_0.png}
    \end{subfigure}
    \begin{subfigure}[t]{0.1\linewidth}
    \includegraphics[width=\linewidth, height=\textheight, keepaspectratio]{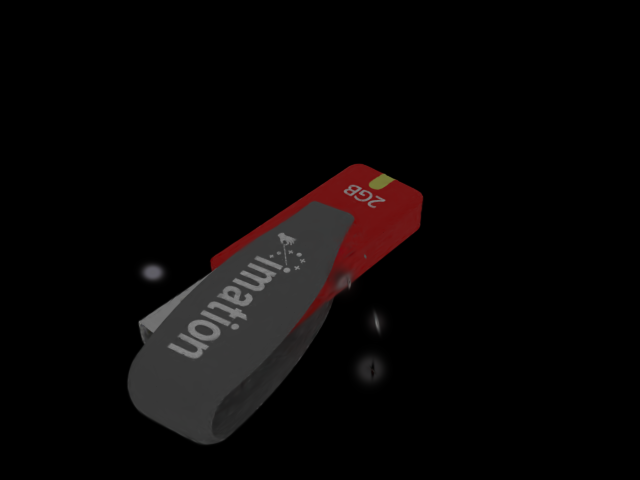}
    \end{subfigure}
    \begin{subfigure}[t]{0.1\linewidth}
    \includegraphics[width=\linewidth, height=\textheight, keepaspectratio]{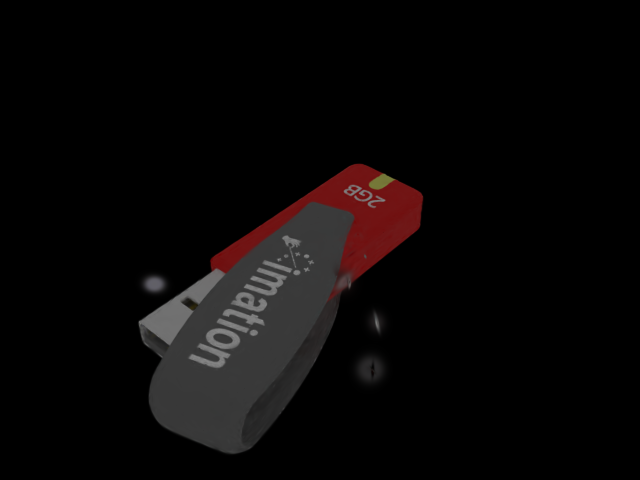}
    \end{subfigure}
    \begin{subfigure}[t]{0.1\linewidth}
    \includegraphics[width=\linewidth, height=\textheight, keepaspectratio]{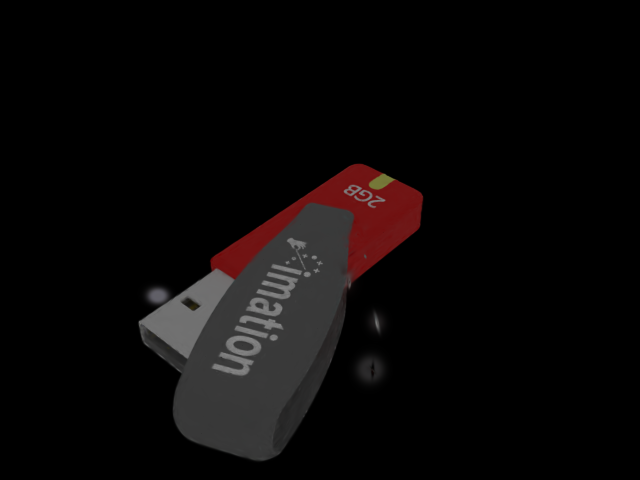}
    \end{subfigure}
    \begin{subfigure}[t]{0.1\linewidth}
    \includegraphics[width=\linewidth, height=\textheight, keepaspectratio]{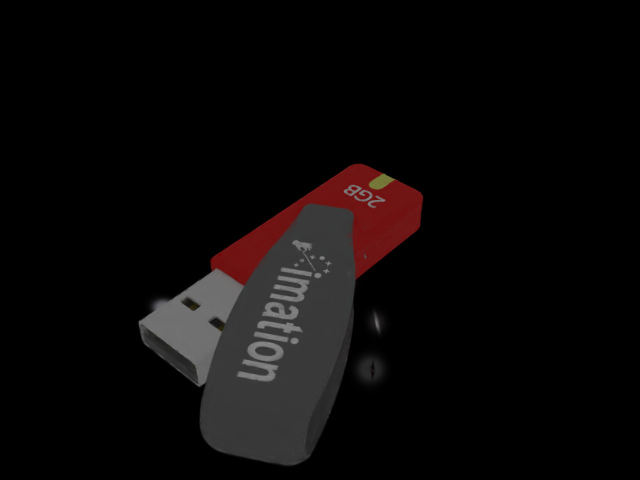}
    \end{subfigure}
    \begin{subfigure}[t]{0.1\linewidth}
    \includegraphics[width=\linewidth, height=\textheight, keepaspectratio]{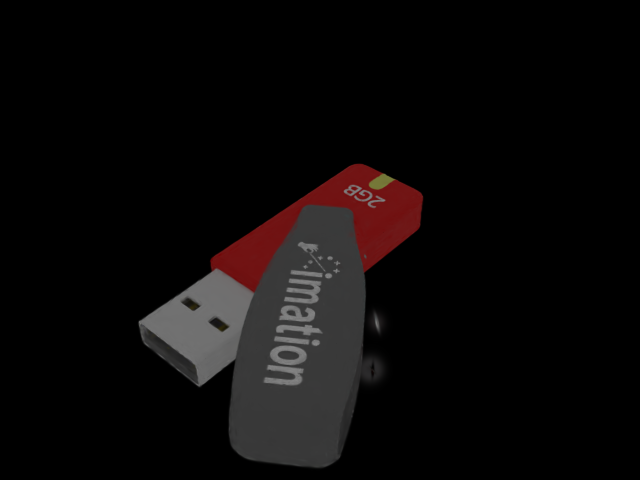}
    \end{subfigure}
    \begin{subfigure}[t]{0.1\linewidth}
    \includegraphics[width=\linewidth, height=\textheight, keepaspectratio]{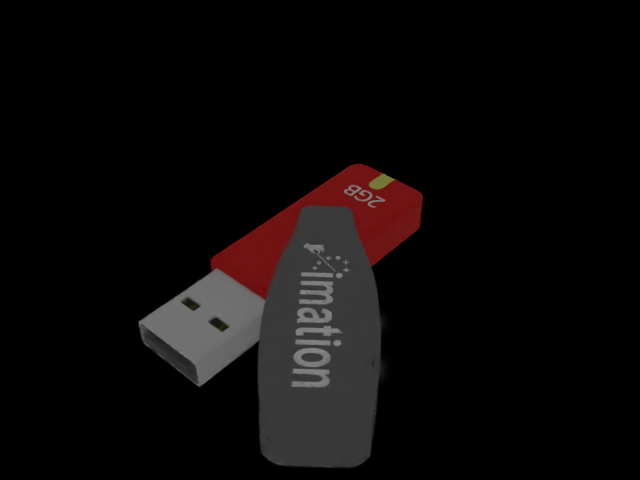}
    \end{subfigure}
    \begin{subfigure}[t]{0.1\linewidth}
    \includegraphics[width=\linewidth, height=\textheight, keepaspectratio]{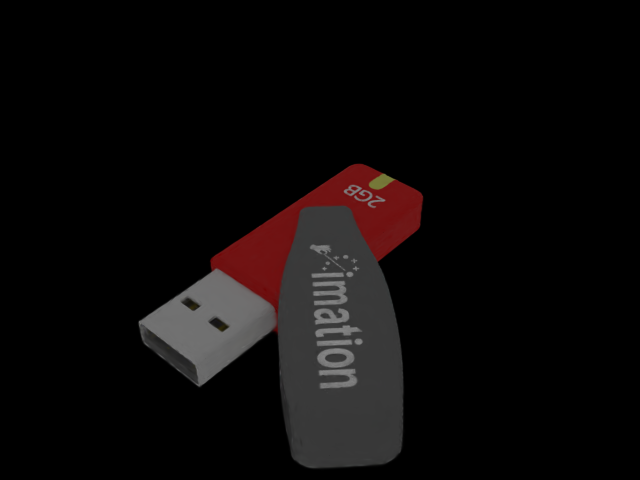}
    \end{subfigure}

    \vspace{0.1cm}
    \begin{subfigure}[t]{0.1\linewidth}
    \includegraphics[width=\linewidth, height=\textheight, keepaspectratio]{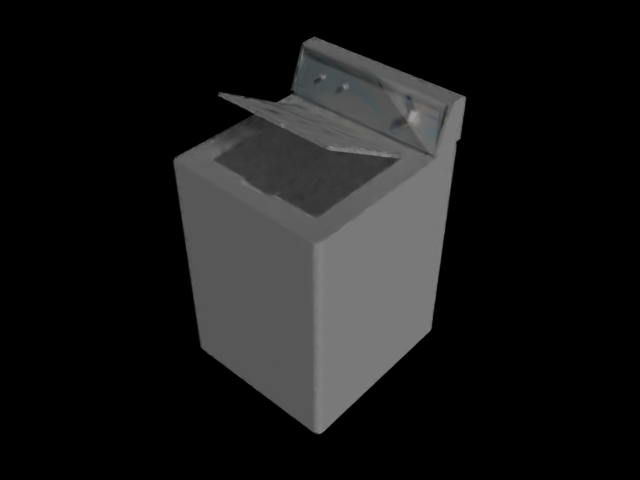}
    \end{subfigure}
    \begin{subfigure}[t]{0.1\linewidth}
    \includegraphics[width=\linewidth, height=\textheight, keepaspectratio]{media/washer/rendered_image_0.png}
    \end{subfigure}
    \begin{subfigure}[t]{0.1\linewidth}
    \includegraphics[width=\linewidth, height=\textheight, keepaspectratio]{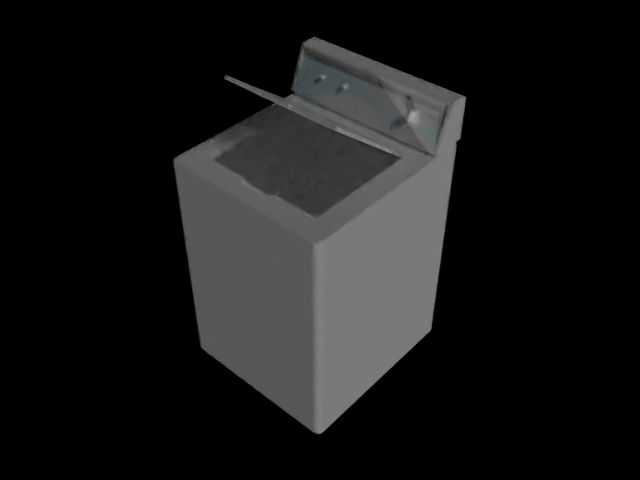}
    \end{subfigure}
    \begin{subfigure}[t]{0.1\linewidth}
    \includegraphics[width=\linewidth, height=\textheight, keepaspectratio]{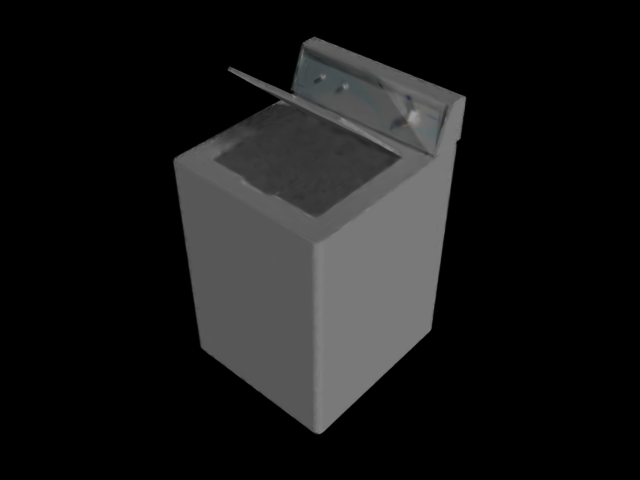}
    \end{subfigure}
    \begin{subfigure}[t]{0.1\linewidth}
    \includegraphics[width=\linewidth, height=\textheight, keepaspectratio]{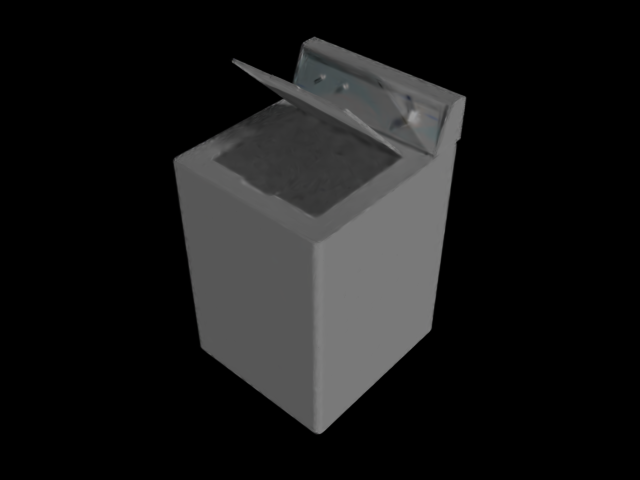}
    \end{subfigure}
    \begin{subfigure}[t]{0.1\linewidth}
    \includegraphics[width=\linewidth, height=\textheight, keepaspectratio]{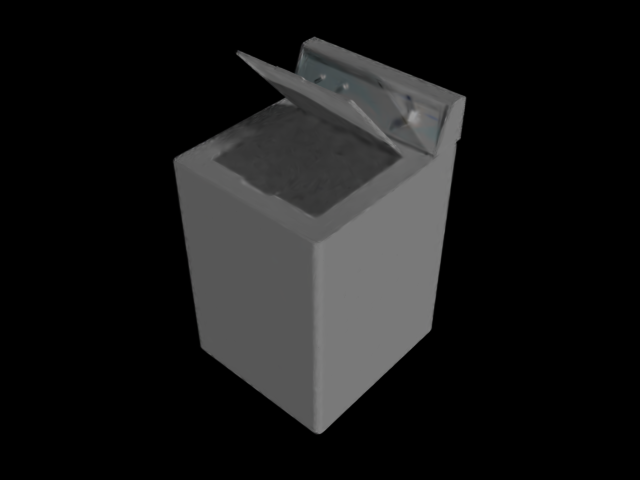}
    \end{subfigure}
    \begin{subfigure}[t]{0.1\linewidth}
    \includegraphics[width=\linewidth, height=\textheight, keepaspectratio]{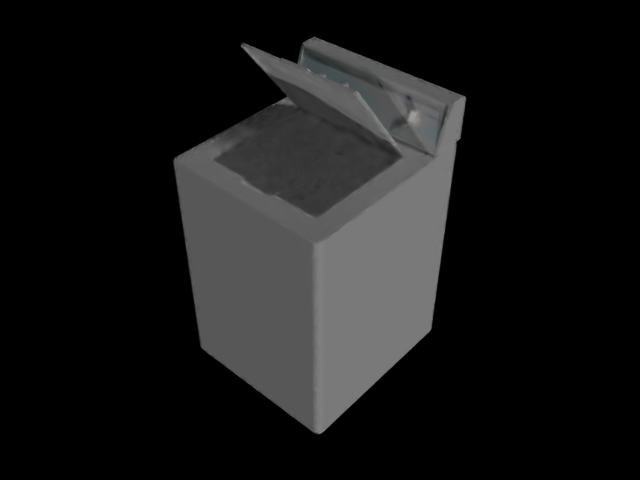}
    \end{subfigure}
    \begin{subfigure}[t]{0.1\linewidth}
    \includegraphics[width=\linewidth, height=\textheight, keepaspectratio]{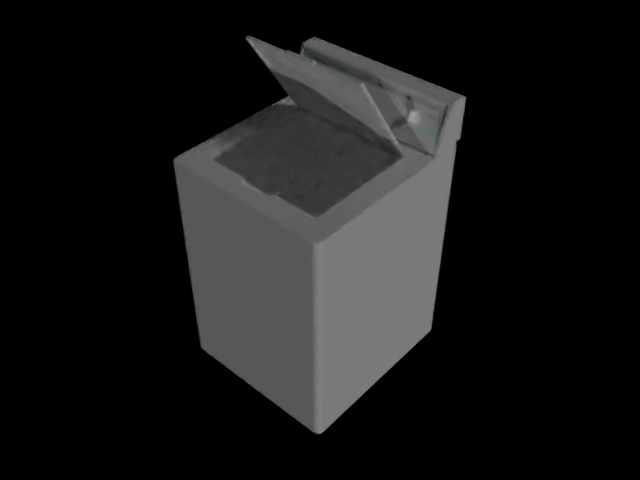}
    \end{subfigure}
    \begin{subfigure}[t]{0.1\linewidth}
    \includegraphics[width=\linewidth, height=\textheight, keepaspectratio]{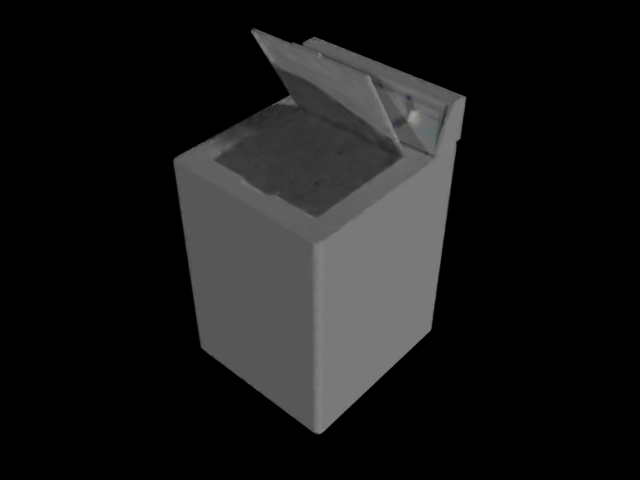}
    \end{subfigure}
    
    \caption{Renderings of the articulated objects within the PARIS dataset while being swept from the learned starting to ending joint parameter. Almost all objects have good qualitative results and good articulation recovery. The fridge presented difficulty in training the base splat - the interior surface being difficult to see into results in noisy splat means, which thus results in a subpar surface approximation. Despite this, Splatart generally recovers the articulation well. The other objects all show excellent articulation estimation results.}
    \label{fig:picwall_paris}
\end{figure*}
\FloatBarrier

\subsection{Joint Estimation}

Previous works such as DTA \cite{weng2024neural} have included the articulation estimate as a part of the primary optimization process as a way to aid in collision checking or other regularization. In contrast, we intentionally sought to remove any aspect of the articulation estimation from the geometry reconstruction and to make joint parameter estimation its own step seperate from the splat and part pose learning processes. This approach allows for greater flexibility in joint specification - more complex articulations such as screws, spherical joints, or kinematic loops can be modeled separately from the reconstruction. Belief space methods can also be leveraged for safety-critical industries. In particular however, we are inspired by previous works that utilized fiducial markers or other pose based estimation to infer kinematic trees. Works such as Sturm et al. \cite{sturm2011probabilistic} have provided pose based solutions for articulated objects that would be advantageous to integrate into modern learning based architectures.

For this work, SPLATART follows a relatively simple articulation estimation procedure. For every pair of part poses $P_{j,k} = \{P^t_j, P^t_k\}_{j\neq k}$ in the canonical scene (which is typically set to $t=0$), we estimate both a revolute and a prismatic joint between them. SPLATART revolute joints are parameterized by $\mathcal{R} = \{c \in \mathbb{R}^3, a \in \mathbb{R}^3, r \in SE(3)\}$ where $c$ is a translation to the rotation center, $a$ is the axis of rotation, and $r$ is a fixed transform from the center, netting a total of twelve (12) degrees of freedom. Prismatic joints are parameterized by $\mathcal{T} = \{ c \in SE(3), a \in \mathbb{R}^3\}$, where $c$ is a fixed transformation, and $a$ is the axis of translation. Both a prismatic and a revolute joint are predicted for each $P_{j,k}$, along with a configuration vector $v$ containing the joint parameters for each scene. These are directly estimated via gradient descent against an ADD matching score loss commonly used for pose estimation \cite{Hinterstoisser_add_matching}. The transforms for both joint candidates are evaluated by transforming the gaussian splat means of the source part $P_j$, and determining the average distance to its corresponding splat mean at $P_k$'s pose as determined previously. If the average distance is below a user-specified threshold $\epsilon$,then the joint is accepted into the graph. Otherwise it is discarded. If both hypothetical joints pass the threshold, then the better performing one is retained.

\begin{table*}[t]
\renewcommand{\arraystretch}{1.0}
\centering
\caption{Quantitative results for Paris dataset objects}
\begin{tabular}{ccccccccccccc}
\toprule
\textbf{Metrics} & \textbf{Methods} & Foldchair & Fridge & Laptop & Oven & Scissor & Stapler & USB & Washer & Blade  & Storage & All \\
\midrule
 Axis Angle (\textdegree) & PARIS & 8.08 & 9.15 & 0.02 & 0.04 & 3.82 & 39.73 & 0.13 & 25.36 & 15.38 & 0.03 & 10.17 \\
 & DTA & 0.03 & 0.07 & 0.06 & 0.22 & 0.11 & 0.06 & 0.11 & 0.43 & 0.27 & 0.06 & 0.14 \\
& SPLATART & 0.63 & 3.46 & 0.35 & 0.24 & 1.09 & 1.68 & 4.61 & 0.46 & 0.43 & 0.33 & 1.33\\
\midrule
 Axis Pos (0.1m) & PARIS &  0.45 & 0.38 & 0.00 & 0.00 & 2.10 & 2.27 & 2.36 & 1.50 & - & - & 1.13 \\
 & DTA & 0.01 & 0.01 & 0.00 & 0.01 & 0.02 & 0.01 & 0.00 & 0.01 & - & - & 0.01 \\
& SPLATART & 0.17 & 0.66 & 0.49 & 0.79 & 0.27 & 0.62 & 0.48 & 0.49 & - & - & 0.50 \\
\midrule

 Part Motion (\textdegree, m) & PARIS & 131.66 & 24.58 & 0.03 & 0.03 & 120.70 & 110.80 & 64.85 & 60.35 & 0.34 & 0.30 & 51.36 \\
 & DTA & 0.16 & 0.09 & 0.08 & 0.11 & 0.15 & 0.05 & 0.11 & 0.25 & 0.00 & 0.00 & 0.10 \\
& SPLATART & 0.02 & 0.15 & 0.05 & 0.06 & 0.01 & 0.09 & 0.01 & 0.04 & 0.00 & .04 & 0.05 \\
\midrule
 CD-s (mm) & PARIS & 33.79 & 3.05 & 0.25 & 2.52 & 39.07 & 41.64 & 2.64 & 10.32 & 46.90 & 9.18 & 18.94 \\
 & DTA & 0.18 & 0.60 & 0.32 & 4.66 & 0.40 & 2.65 & 2.19 & 4.80 & 0.55 & 4.69 & 2.10 \\
& SPLATART & 0.03 & 0.23 & 0.04 & 0.48 & 0.06 & 0.15 & 0.25 & 0.88 & 0.04 & 0.77 & 0.29 \\
\midrule
 CD-m (mm) & PARIS & 8.99 & 7.76 & 0.21 & 28.70 & 46.64 & 19.27 & 5.32 & 178.43 & 25.21 & 76.69 & 39.72 \\
 & DTA & 0.15 & 0.27 & 0.16 & 0.47 & 0.41 & 2.27 & 1.34 & 0.36 & 1.50 & 0.37 & 0.73 \\
& SPLATART & 0.03 & 20.35 & 0.03 & 0.24 & 0.06 & 0.18 & 0.09 & 0.56 & 0.05 & 0.70 & 2.23 \\
\midrule
 CD-w (mm) & PARIS & 1.80 & 2.92 & 0.30 & 11.73 & 10.49 & 3.58 & 2.00 & 24.38 & 0.60 & 8.57 & 6.64 \\
 & DTA & 0.27 & 0.70 & 0.35 & 4.18 & 0.43 & 2.19 & 1.18 & 4.74 & 0.36 & 3.99 & 1.84 \\
& SPLATART & 0.03 & 5.43 & 0.04 & 0.47 & 0.06 & 0.15 & 0.15 & 0.87 & 0.03 & 0.55 & 0.78 \\
\bottomrule
\bottomrule
\end{tabular}
\label{tab:paris_quantitative}
\caption{Quantitative results for the relative joint states. }
\end{table*}

\subsection{Configurable Rendering}

 Subsequent to estimating the full joint connectivity graph, we can perform configurable rendering with the learned gaussian splat. Starting from a user-designated root node part, a depth-first search is performed over the joint connectivity graph to produce the final kinematic tree. This tree-based approach is to ensure that no kinematic loops are created in our joint representations - although such a structure is in theory detectable and solvable, we leave loopy objects for future research efforts. With the obtained set of joints, a configuration vector can then either be instantiated to match an example from the training set, or it can be provided by the user for some downstream robot task. During rendering, the joint parameter tree is once again traversed to generate out the list of part poses $\{P_j\}_{j=1}^{N_p}$, which are applied to the previously learned gaussian parameters $\mu$ and $q$ for rasterization.

\begin{table}
\centering
\begin{tabular}{lll}
Joint Num & Type      & Error (deg, mm)   \\
\midrule
1         & Revolute  & 0.87              \\
2         & Revolute  & 0.39              \\
3         & Revolute  & 0.10              \\
4         & Revolute  & 0.03              \\
5         & Revolute  & 0.39              \\
6         & Revolute  & 0.09              \\
7         & Revolute  & 0.99              \\
8         & Prismatic & 1.17              \\
9         & Prismatic & 0.53             
\end{tabular}
\caption{Joint estimation errors for the Panda arm example shown in figure \ref{fig:panda_close}}
\label{tab:panda_joints}
\end{table}

\section{Results}
\subsection{Paris Two-Part Object Dataset}
PARIS\cite{liu2023paris} is a dataset comprised of a collection of single degree-of-freedom articulated objects drawn from both the PartNet-Mobility dataset \cite{xiang2020sapien}, and from Multiscan\cite{mao2022multiscan}. Every object is presented in a start and end state, with only one part of the object having any motion - the other remains fixed within the world frame. Each object has 100 random spherical RGB images and object masks. For our method we augmented this data with renderings of the part masks from the Sapien renderer.
\subsubsection{Paris Experiment}
To demonstrate the base capabilities of SPLATART, we apply it to the segmented Paris dataset and present comparisons to Paris\cite{liu2023paris} amd DTA \cite{weng2024neural} in table \ref{tab:paris_quantitative}. We additionally show qualitative renderings of the PartNet-Mobility objects in Figure \ref{fig:picwall_paris} to demonstrate the effectiveness of SPLATART's cross scene pose and geometry estimation approach. In this experiment, the base splats were trained for 20,000 iterations on the Splatfacto model in NerfStudio \cite{tancik2023nerfstudio}, which has been augmented with an accumulation loss to aid in convergence speed. The segmentation learning and pose learning steps were run for 20 epochs per scene, and the learning rates were kept the same throughout all trials. All models were trained on an RTX 3090 GPU which took approximately 20 minutes from end-to-end. 

SPLATART performs exceptionally well at reconstructing the part geometry. SPLATART's access to the part segmentation mask assists greatly in this, just as DTA's access to depth assists its performance. The exception to this is the fridge, where the splat training process has difficulty training well for the interior geometry of the object. SPLATART also performs very well at part motion estimation, excepting a few cases where the moving part had exceptionally poor reconstruction. SPLATART's articulation model is possibly differently parameterized than other articulated efforts, which may explain the different axis position metrics versus the baselines, but the recovered part motion values remain highly accurate.

\begin{figure*}[t]
    \begin{subfigure}[t]{0.5\linewidth}
    \includegraphics[width=\linewidth, height=\textheight, keepaspectratio]{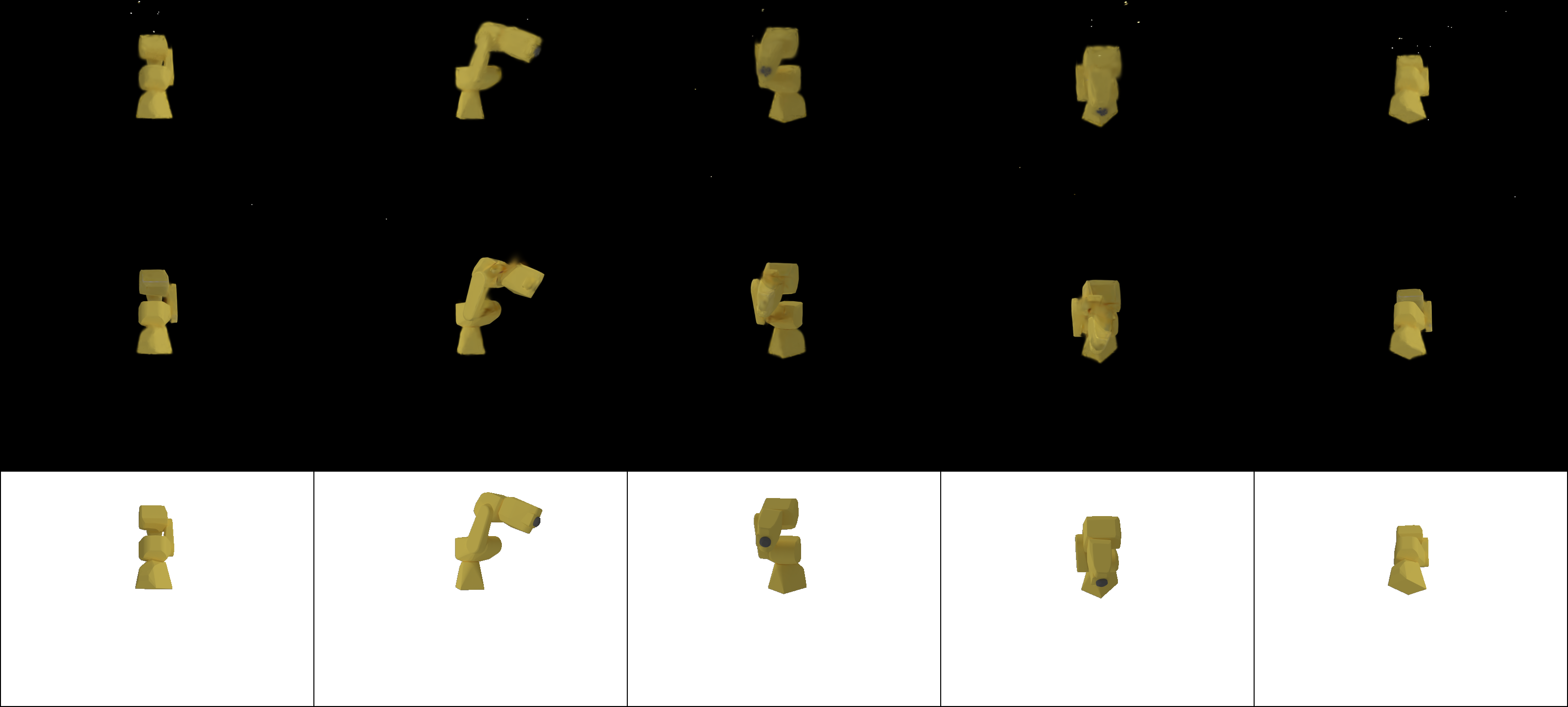}
    \end{subfigure}
    \begin{subfigure}[t]{0.5\linewidth}
    \includegraphics[width=\linewidth, height=\textheight, keepaspectratio]{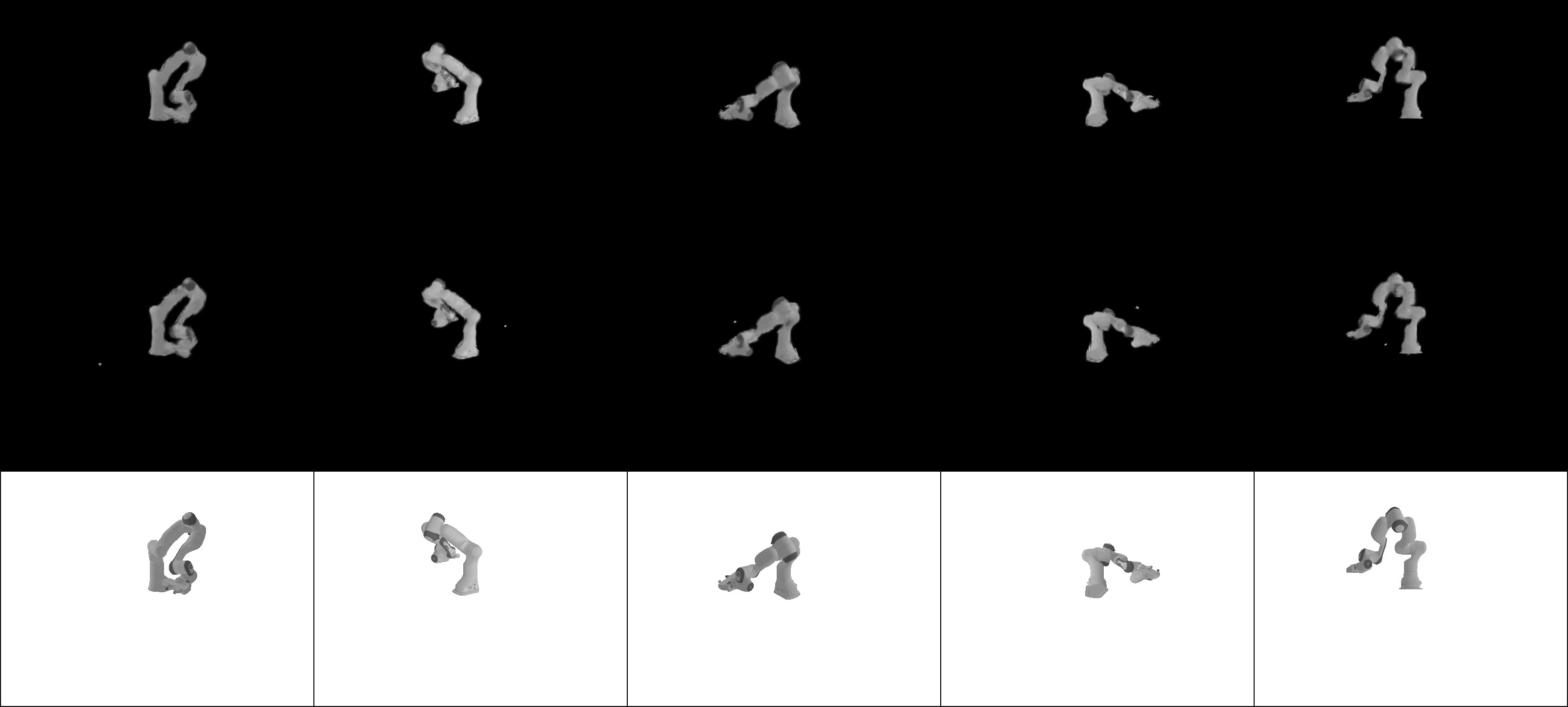}
    \end{subfigure}

    \vspace{0.1cm}
    \begin{subfigure}[t]{0.5\linewidth}
    \includegraphics[width=\linewidth, height=\textheight, keepaspectratio]{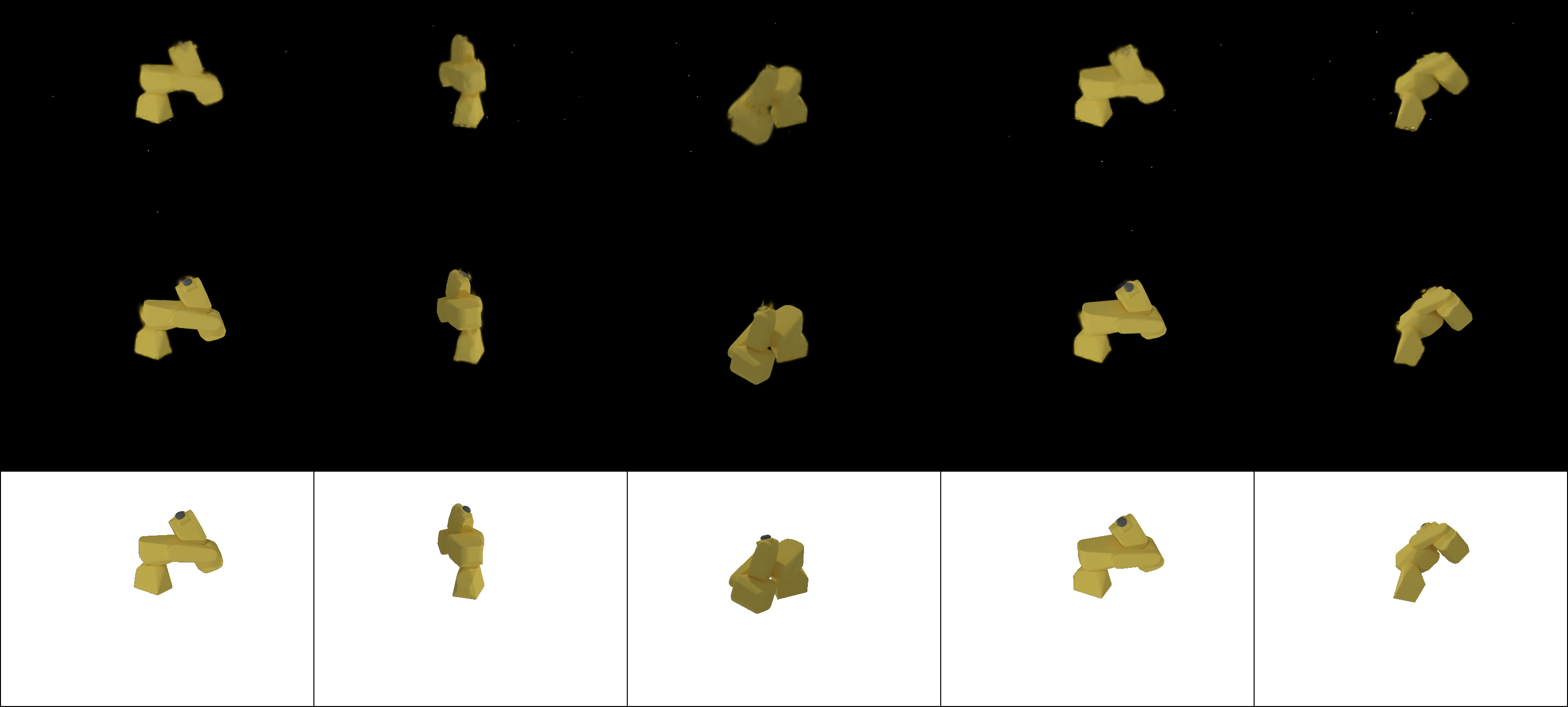}
    \end{subfigure}
    \begin{subfigure}[t]{0.5\linewidth}
    \includegraphics[width=\linewidth, height=\textheight, keepaspectratio]{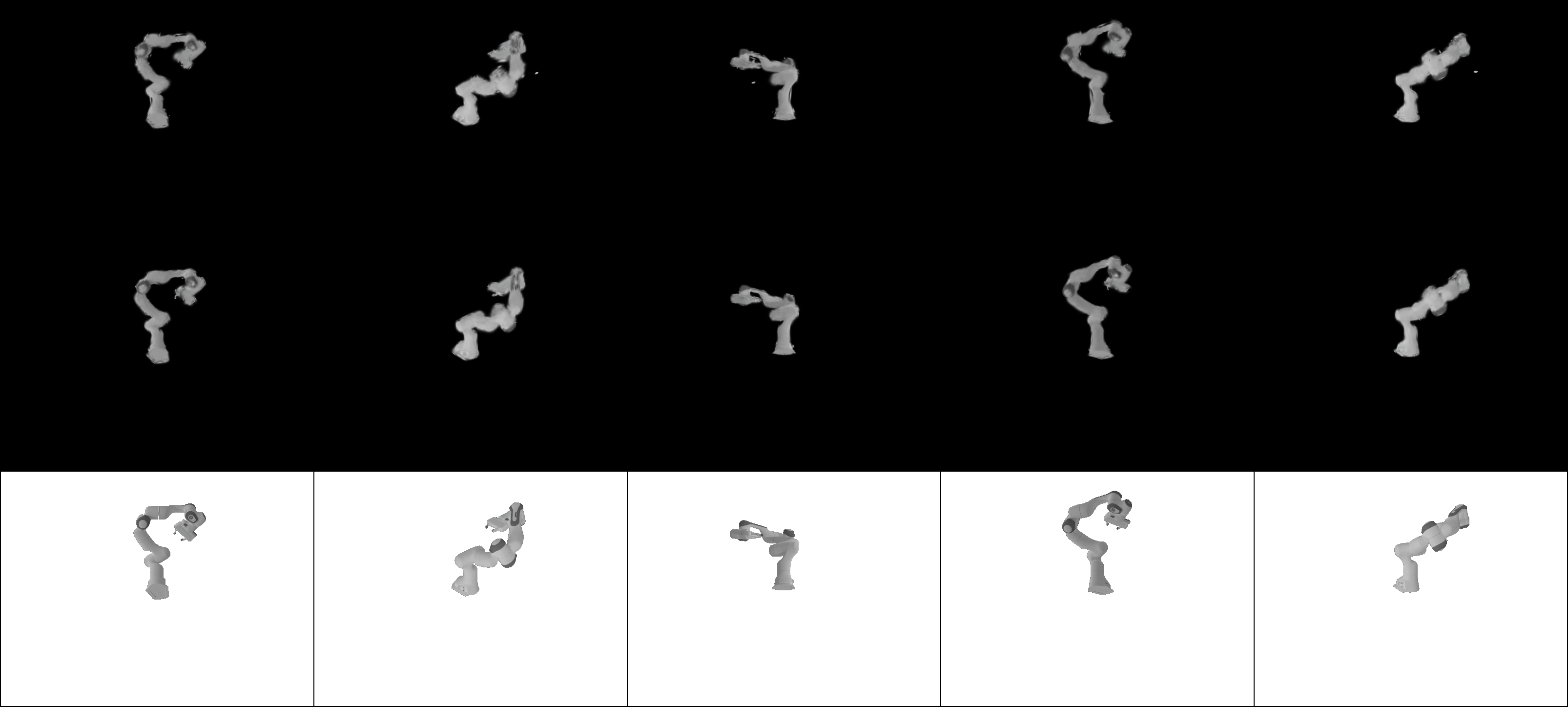}
    \end{subfigure}
\caption{Renderings of serial chain manipulators from the URDF files Dataset showing SPLATART's ability to extract poses and joints even with large numbers of articulations. For each quadrant of images, the bottom images (with the white background) are the ground truth images. The top row of each row of three corresponds to the splats trained on data for $t=0$, and the middle row is for data on $t=1$. The top set of three rows are the poses for time $t=0$, and the bottom set are for time $t=1$.}
    \label{fig:picwall_robots}
\end{figure*}

\begin{figure*}[t]
    \begin{subfigure}[t]{0.33\linewidth}
    \includegraphics[width=\linewidth, height=\textheight, keepaspectratio]{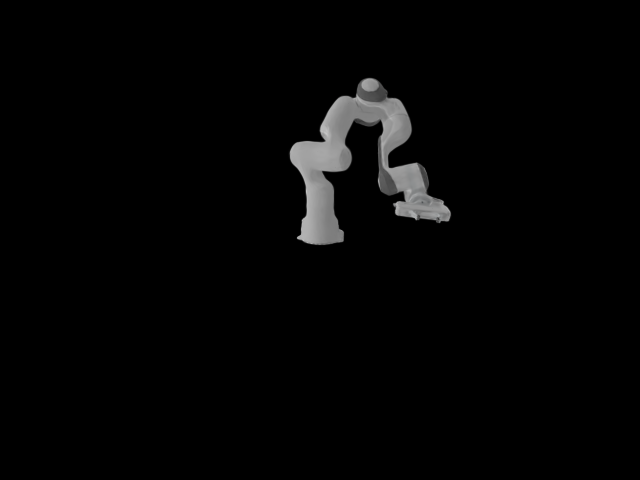}
    \end{subfigure}
    \begin{subfigure}[t]{0.33\linewidth}
    \includegraphics[width=\linewidth, height=\textheight, keepaspectratio]{media/panda/rendered_image_0.png}
    \end{subfigure}
    \begin{subfigure}[t]{0.33\linewidth}
    \includegraphics[width=\linewidth, height=\textheight, keepaspectratio]{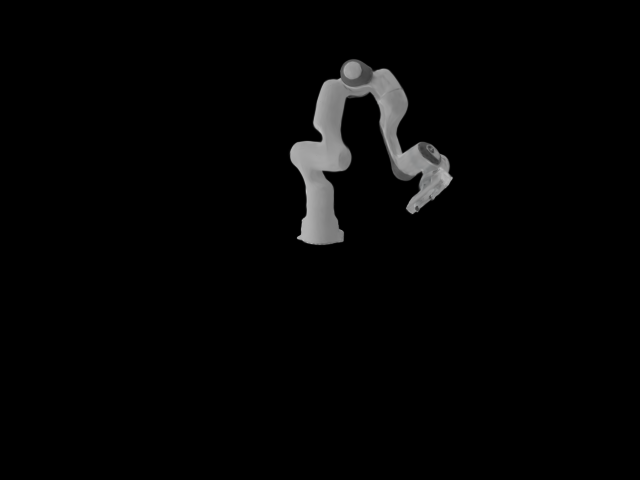}
    \end{subfigure}
    \begin{subfigure}[t]{0.33\linewidth}
    \includegraphics[width=\linewidth, height=\textheight, keepaspectratio]{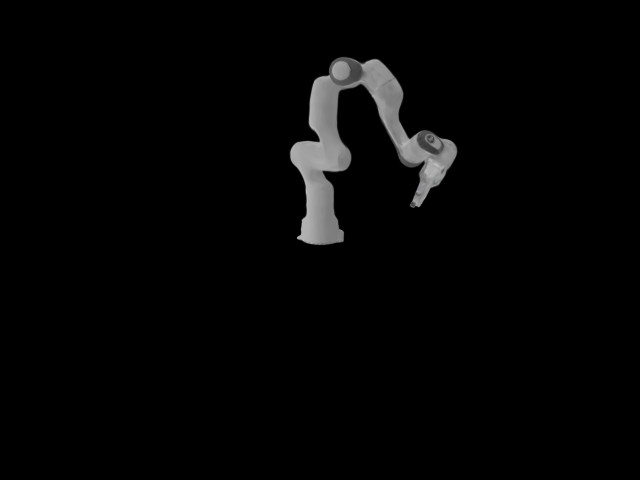}
    \end{subfigure}
    \begin{subfigure}[t]{0.33\linewidth}
    \includegraphics[width=\linewidth, height=\textheight, keepaspectratio]{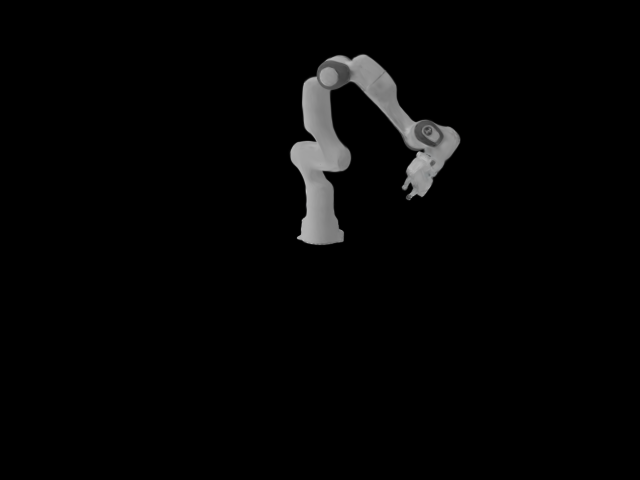}
    \end{subfigure}
    \begin{subfigure}[t]{0.33\linewidth}
    \includegraphics[width=\linewidth, height=\textheight, keepaspectratio]{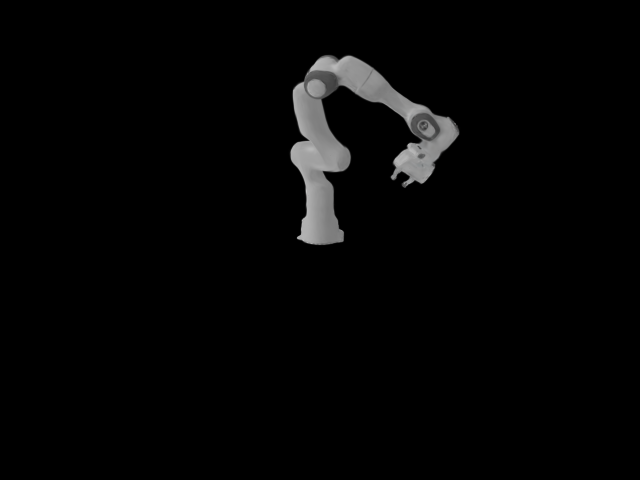}
    \end{subfigure}
    \begin{subfigure}[t]{0.33\linewidth}
    \includegraphics[width=\linewidth, height=\textheight, keepaspectratio]{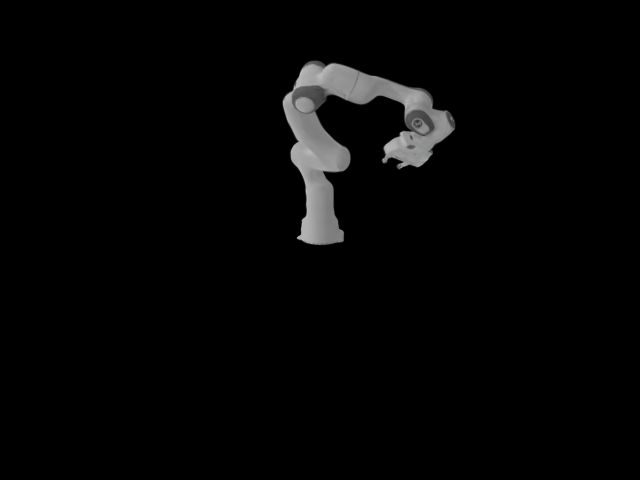}
    \end{subfigure}
    \begin{subfigure}[t]{0.33\linewidth}
    \includegraphics[width=\linewidth, height=\textheight, keepaspectratio]{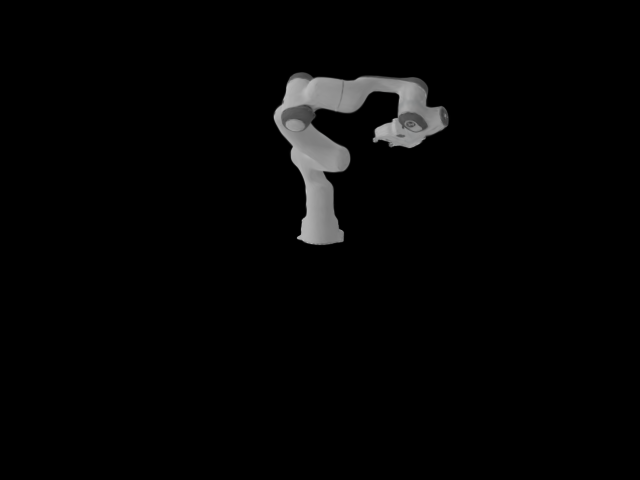}
    \end{subfigure}
    \begin{subfigure}[t]{0.33\linewidth}
    \includegraphics[width=\linewidth, height=\textheight, keepaspectratio]{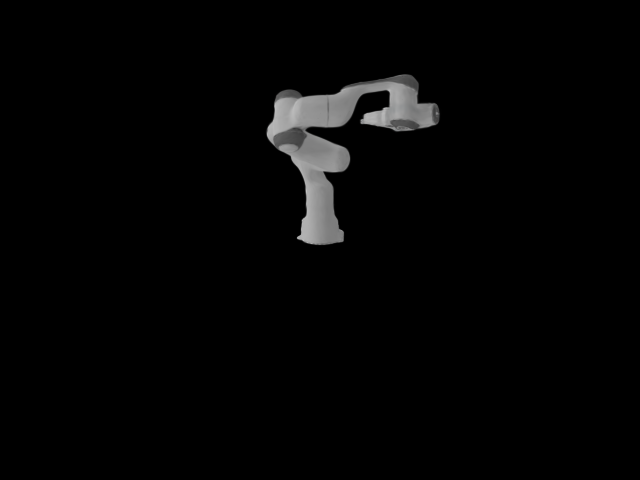}
    \end{subfigure}
    
    \caption{Renderings of the Panda arm as it moves through intermediate states that were unobserved during training. The top left image represents the robot's configuration at $t$=0. The bottom right image represents the robot's configuration at $t$=1. Each joint's state is linearly interpolated between the estimated $t=0$ and $t=1$ configurations, with the rendering results being displayed sequentially left to right per row.}
    \label{fig:panda_close}
\end{figure*}

\begin{figure}[H]
    \centering
    \begin{tabular}{ccc}
        {\includegraphics[width=0.13\textwidth]{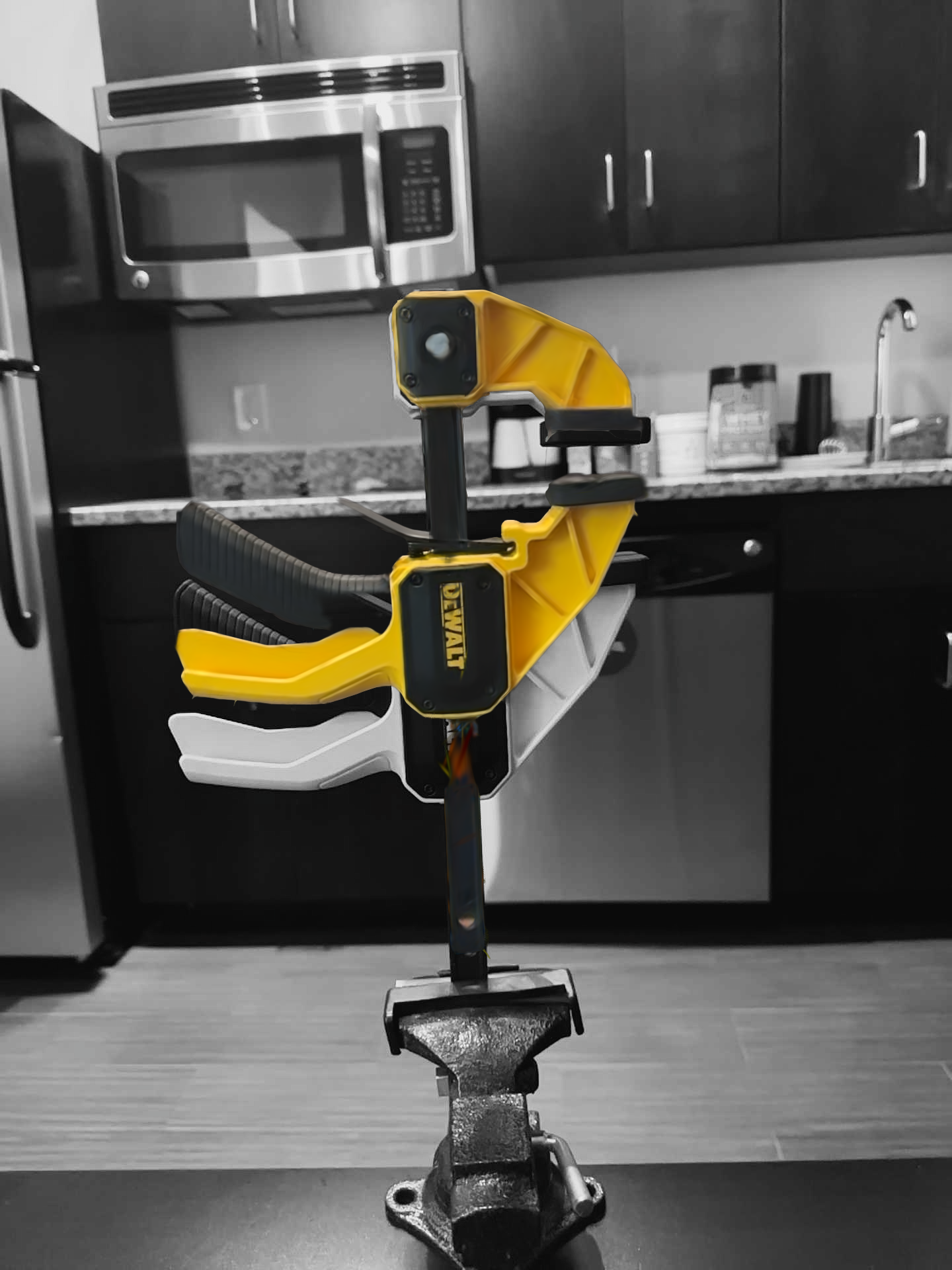}} &
        {\includegraphics[width=0.13\textwidth]{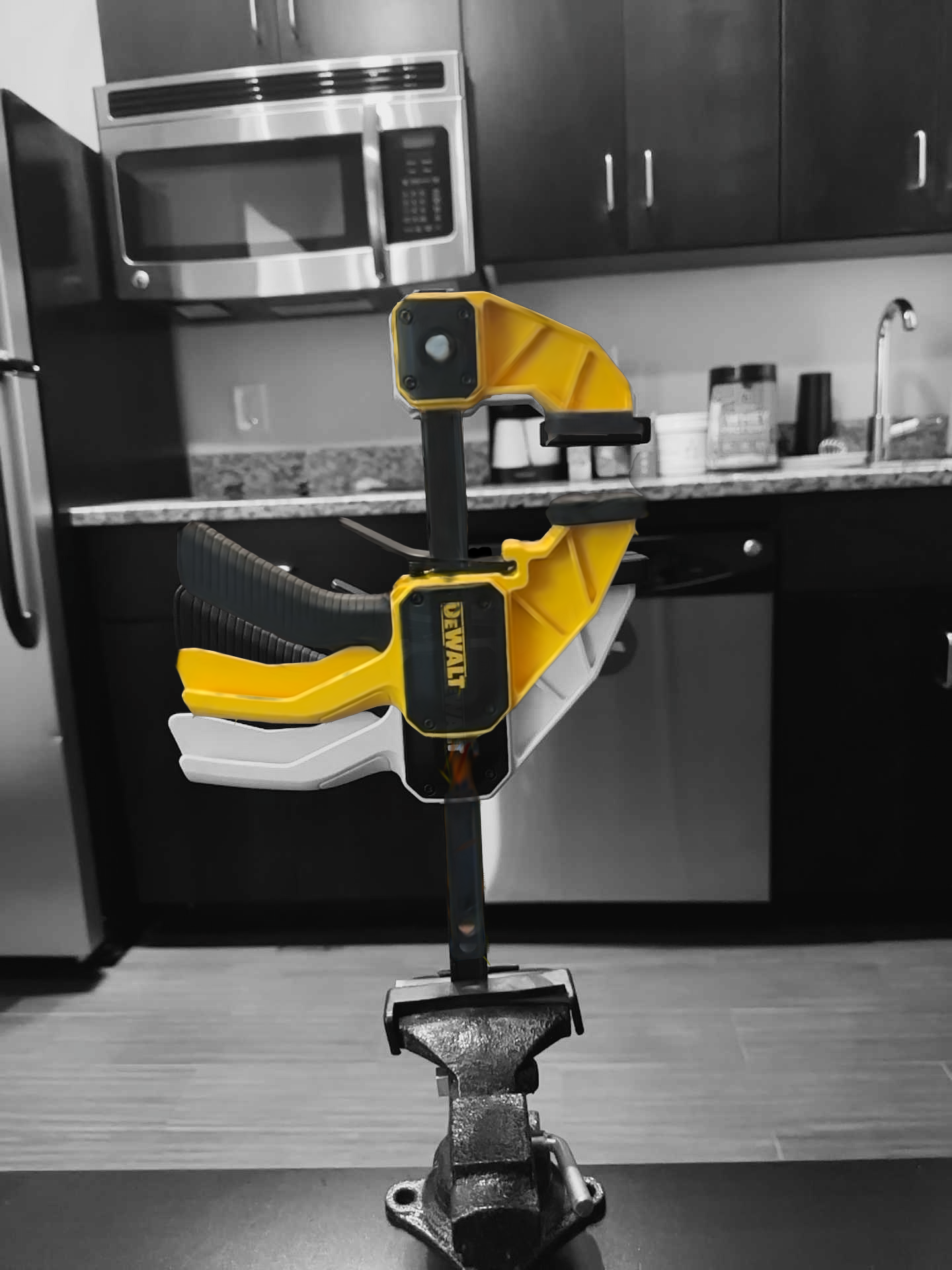}} &
        {\includegraphics[width=0.13\textwidth]{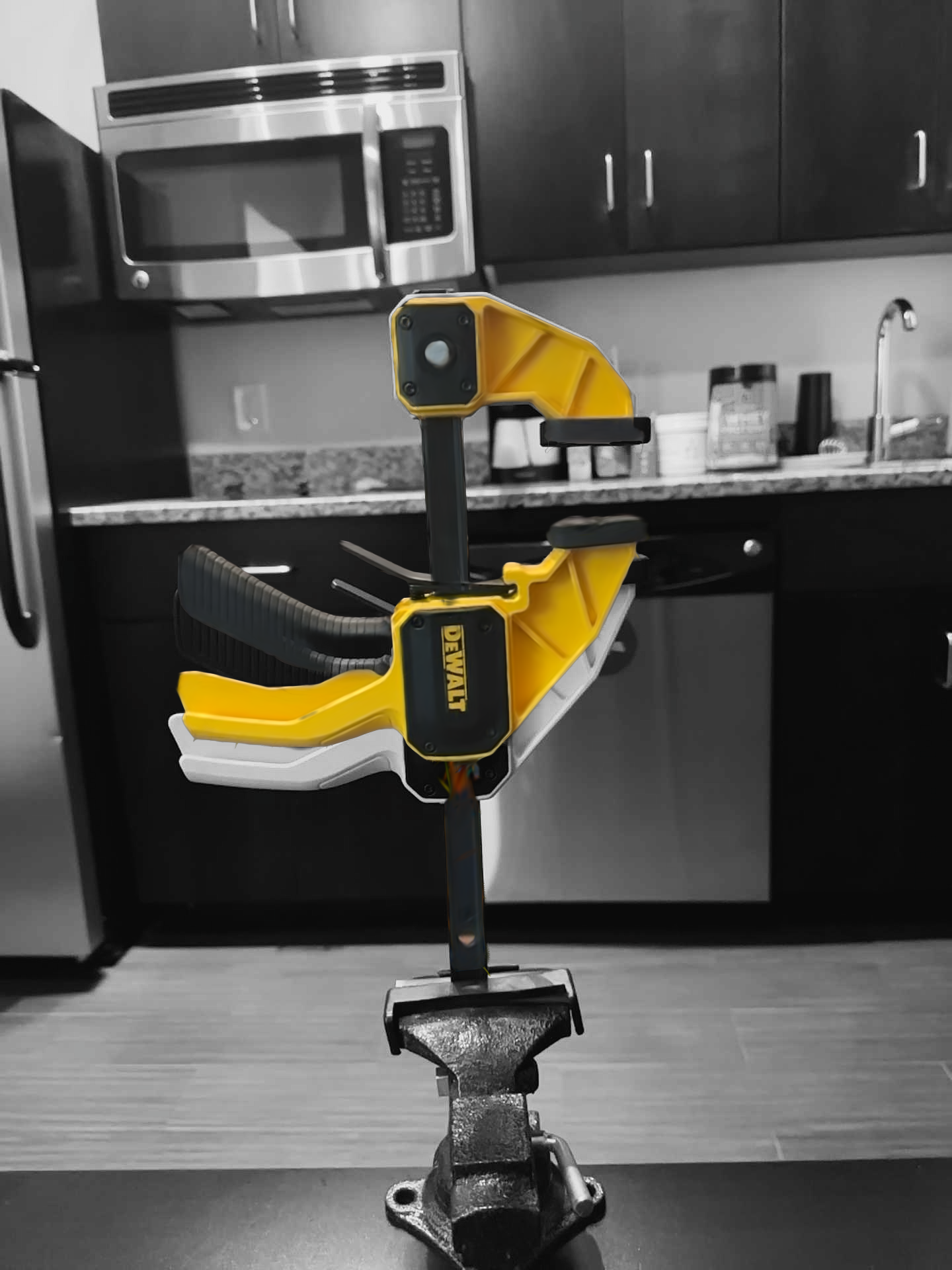}} \\
        {\includegraphics[width=0.13\textwidth]{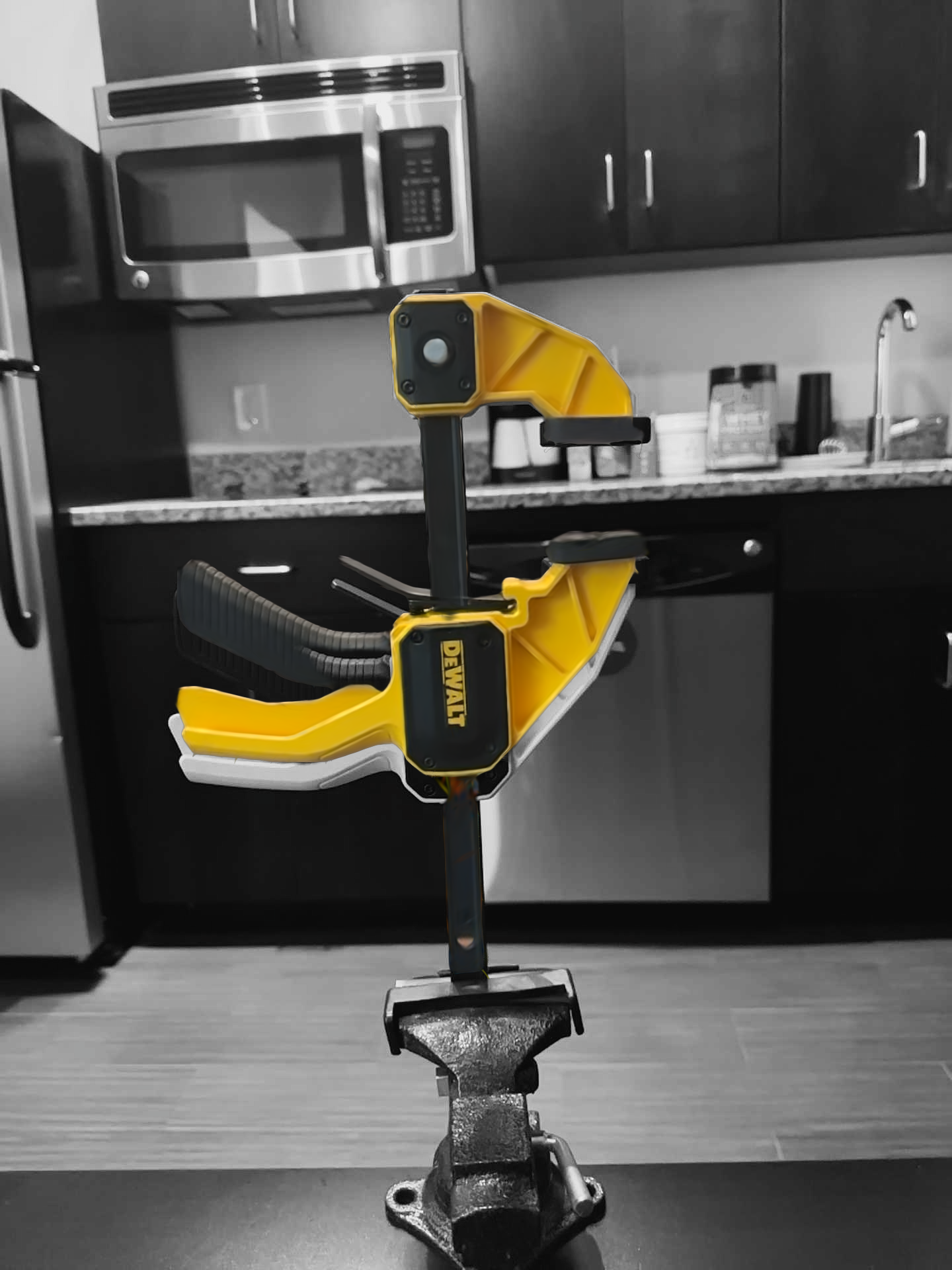}} &
        {\includegraphics[width=0.13\textwidth]{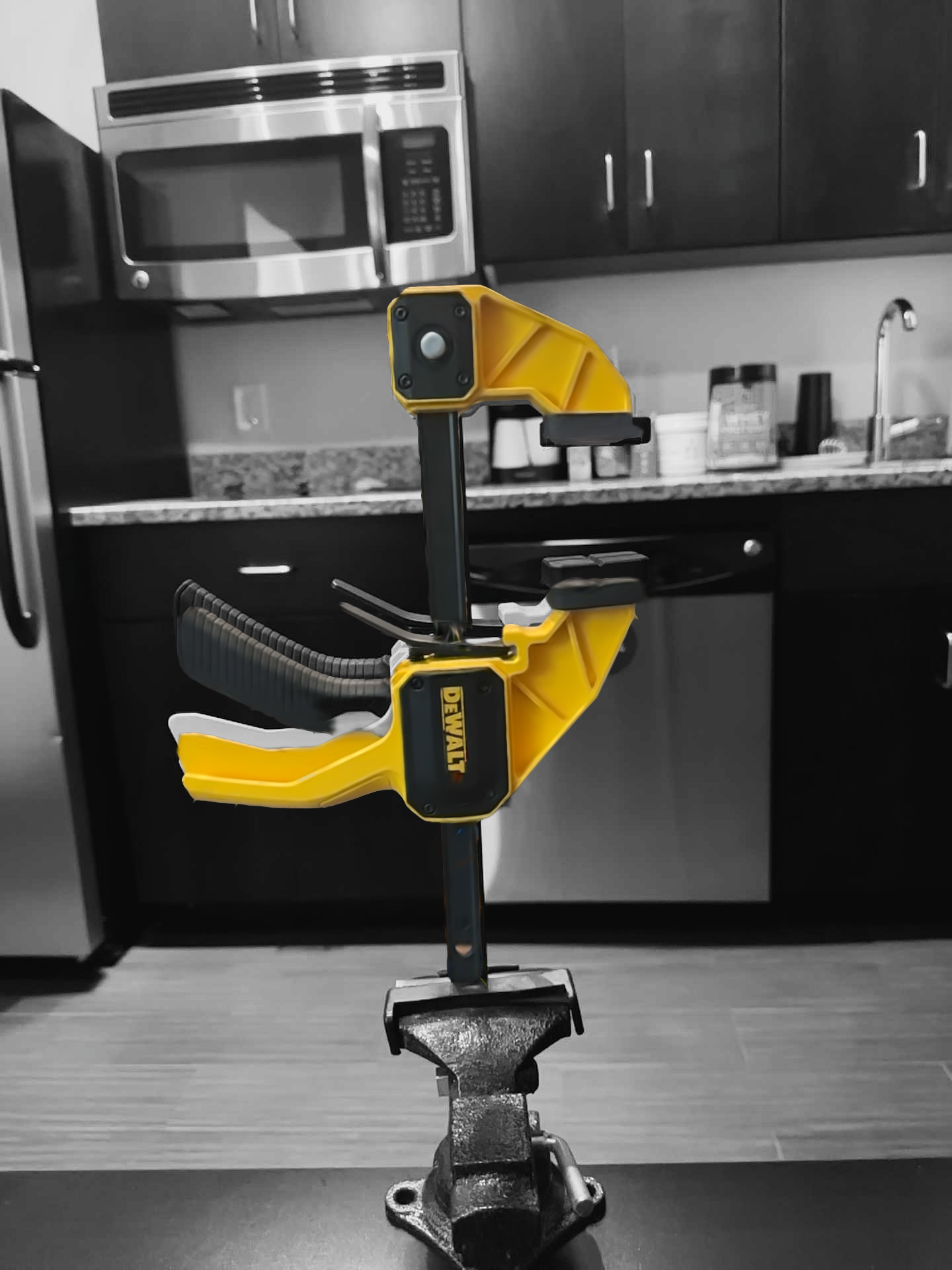}} &
        {\includegraphics[width=0.13\textwidth]{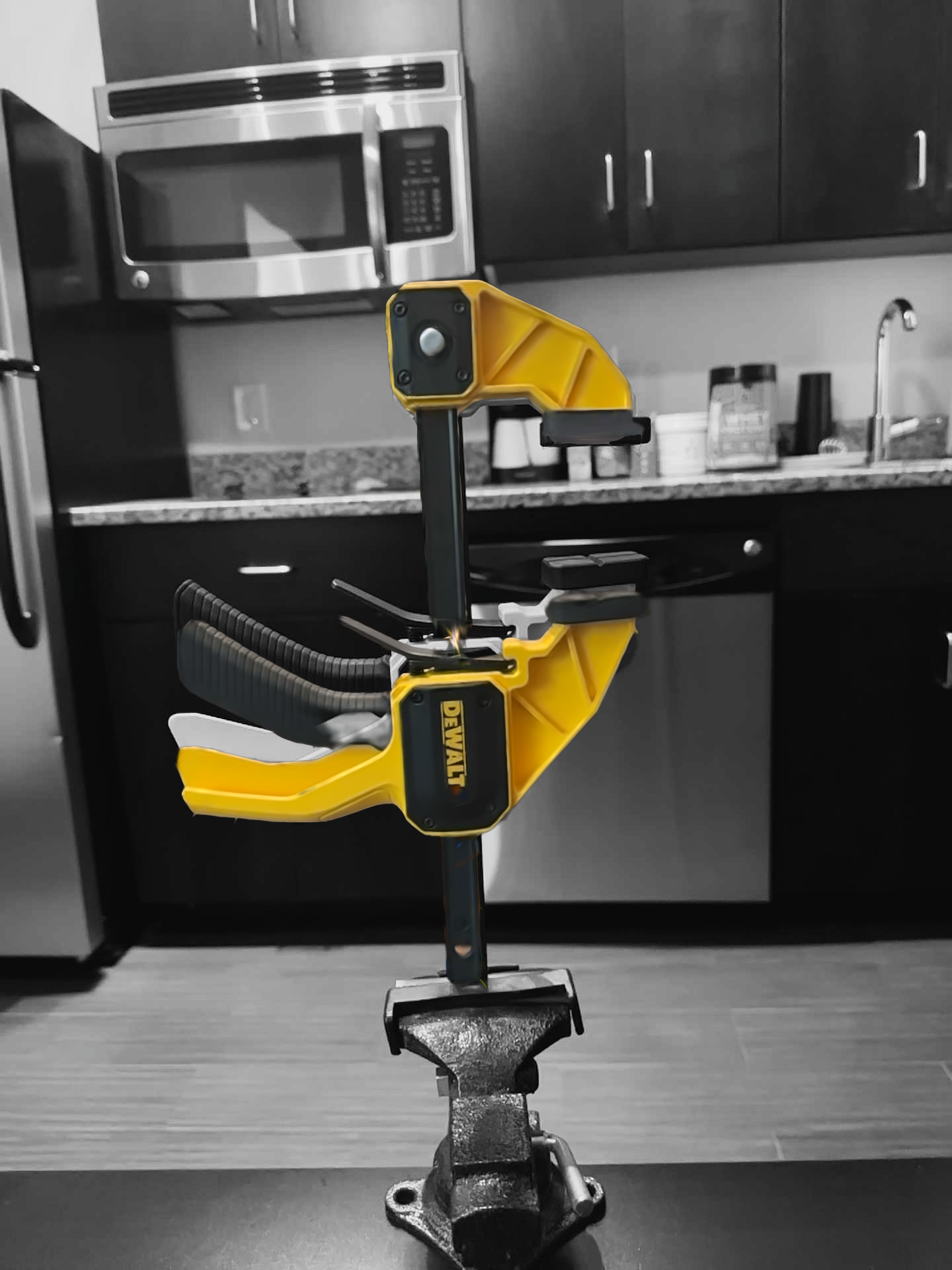}} \\
    \end{tabular}
    \caption{6 novel scenes synthesized with our method, overlaid on the original training image. The clamp progressively opens from top left to bottom right. This was achieved with only a fraction of the input images having segmentation labels.}
    \label{fig:novelview}
\end{figure}

\subsection{Real World Sparse Segmentations}
Though we have shown the proposed method to be effective on synthetic datasets, applying a method that depends on segmentations to real world objects may prove more laborious than desired. Thus, we show examples of a real-world hand tool where the segmentations are only sparsely labeled. Training data for this experiment was collected with a single iPhone 14 Pro's built-in LiDAR and the Record 3D app \cite{record3d}. The segmentation was subsequently performed manually using Roboflow \cite{roboflow_software}. In total, only 30 images were labeled in each of 3 scenes of a trigger clamp (total 90 images). From this, several novel scenes (i.e. novel clamp positions) were rendered. Qualitative results for novel rendering are shown in Figure \ref{fig:novelview} where it can be seen that despite the low prevalence of segmentations, good geometric and kinematic results were obtained.

\subsection{Serial Chain Manipulators}
One of SPLATART's primary goals was to achieve success on objects with larger kinematic tree structures. We show results here on running two synthetic serial chain manipulator models through the SPLATART pipeline - a Franka Panda model, and a Fanuc lr mate 200id. The URDF and model files for these robots are available via the URDF Dataset \cite{URDFDataset}. Both the training RGB images and segmentations were rendered using the Sapien renderer with raytracing enabled \cite{xiang2020sapien}. The renderings shown in Figure \ref{fig:picwall_robots} demonstrate that even with a highly articulated arm with up to seven degrees of freedom, SPLATART is able to obtain the correct poses and geometry for each arm part. Unlike when training for the Paris dataset objects, the robot arm parts were initially centered according to the mean of their gaussian $\mu$ parameters, and a local alignment with ICP (iterative closest point) was utilized to ease the numerical stability on optimization startup. In Figure \ref{fig:panda_close} we show intermediate states of the Franka Panda in more detail, with the intermediate figures representing configuration states not observed during the training process. We additionally show the the estimated joint motion error for the Panda example from Figure \ref{fig:panda_close} in table \ref{tab:panda_joints}. Further renderings and animations will be available on the project's website for inspection.

\section{Conclusion}

In conclusion, SPLATART demonstrates the capability to generate articulated object renderers via gaussian splatting without a prior knowledge of the object structure. It achieves results comprable to previous radiance field based efforts such as PARIS, and additionally is able to generate renderers from real world collected data with a minimal amount of human labelling required. SPLATART additionally demonstrated the ability to extract part pose and articulation models from articulated objects with high depth kinematic trees by showing rendering results for serial chain manipulators. It is important to note that while the addition of segmentations is a drawback of SPLATART's approach, however that is ameliorated by the rise of Segment Anything \cite{kirillov2023segment} based labelling tools. Future work will seek to explore how SPLATART may retain its ability to model complex articulated objects without the need for additional segmentation inputs. Further information on this work and additional images will be posted on its website at https://progress.eecs.umich.edu/projects/splatart/.



\bibliographystyle{ieeetr}
\bibliography{biblio}

\end{document}